\documentclass[10pt,twocolumn,letterpaper]{article}

\usepackage{iccv}              
\usepackage{diagbox}
\definecolor{iccvblue}{rgb}{0.21,0.49,0.74}
\usepackage[pagebackref,breaklinks,colorlinks,allcolors=iccvblue]{hyperref}

\usepackage{graphicx} 
\usepackage{subcaption} 
\usepackage{multirow}

\usepackage{colortbl}
\usepackage{amsmath}
\usepackage{amssymb}
\usepackage{booktabs}
\usepackage{bigstrut}
\usepackage{array}
\usepackage{tabularx}
\usepackage{tabu}
\usepackage{ulem}
\def\paperID{14382} 
\def\confName{ICCV}
\def\confYear{2025}

\title{Efficient Adaptation of Pre-trained Vision Transformer underpinned by Approximately Orthogonal Fine-Tuning Strategy}

\author{Yiting Yang$^1$, Hao Luo$^1$, Yuan Sun$^2$ ,Qingsen Yan$^3$, Haokui Zhang$^3$, Wei Dong$^1$\thanks{Corresponding author. (dongwei156@xauat.edu.cn)}, \\
Guoqing Wang$^2$, Peng Wang$^2$,Yang Yang $^2$, Hengtao Shen$^4$\\
$^{1}$ Xi'an University of Architecture and Technology\\
$^{2}$ University of Electronic Science and Technology of China\\
$^{3}$ Northwestern Polytechnical University\quad$^{4}$ TongJi University\\
}

\begin{document}
\maketitle
\begin{abstract}

A prevalent approach in Parameter-Efficient Fine-Tuning (PEFT) of pre-trained Vision Transformers (ViT) involves freezing the majority of the backbone parameters and solely learning low-rank adaptation weight matrices to accommodate downstream tasks. These low-rank matrices are commonly derived through the multiplication structure of down-projection and up-projection matrices, exemplified by methods such as LoRA and Adapter. In this work, we observe an approximate orthogonality among any two row or column vectors within any weight matrix of the backbone parameters; however, this property is absent in the vectors of the down/up-projection matrices. Approximate orthogonality implies a reduction in the upper bound of the model's generalization error, signifying that the model possesses enhanced generalization capability. If the fine-tuned down/up-projection matrices were to exhibit this same property as the pre-trained backbone matrices, could the generalization capability of fine-tuned ViTs be further augmented? To address this question, we propose an Approximately Orthogonal Fine-Tuning (AOFT) strategy for representing the low-rank weight matrices. This strategy employs a single learnable vector to generate a set of approximately orthogonal vectors, which form the down/up-projection matrices, thereby aligning the properties of these matrices with those of the backbone. Extensive experimental results demonstrate that our method achieves competitive performance across a range of downstream image classification tasks, confirming the efficacy of the enhanced generalization capability embedded in the down/up-projection matrices. Our code is available at link\footnote{\url{https://drive.google.com/file/d/1rg3JYfkmeLGDbRWXspO22wxVspbtnthV/view?usp=drive_link}}.

\end{abstract}    
\vspace{-0.5cm}
\section{Introduction}

To leverage the strengths of pre-trained Vision Transformers (ViT)~\cite{dosovitskiy2020image} for specific downstream applications, Parameter-Efficient Fine-Tuning (PEFT) strategy has gained significant attention. This PEFT~\cite{hu2021lora, jia2022visual, houlsby2019parameter, kopiczko2023vera, dettmers2023qlora} strategy aims to minimize the number of parameters that need to be updated, thereby enhancing computational efficiency and reducing storage requirements. A prevalent approach within PEFT involves freezing the majority of the backbone parameters of the ViT and focusing on learning low-rank adaptation weight matrices to accommodate the nuances of different downstream tasks. Methods such as LoRA~\cite{hu2021lora} (Low-Rank Adaptation) and Adapter~\cite{houlsby2019parameter} have exemplified this approach by deriving low-rank matrices through the multiplication structure of down- and up-projection matrices. These matrices act as intermediary layers that facilitate the adaptation of the frozen backbone to new tasks, while keeping the number of learnable parameters to a minimum.

\begin{figure}[!tb]
    \centering
     \begin{subfigure}[b]{0.22\textwidth}
        \includegraphics[width=\textwidth]{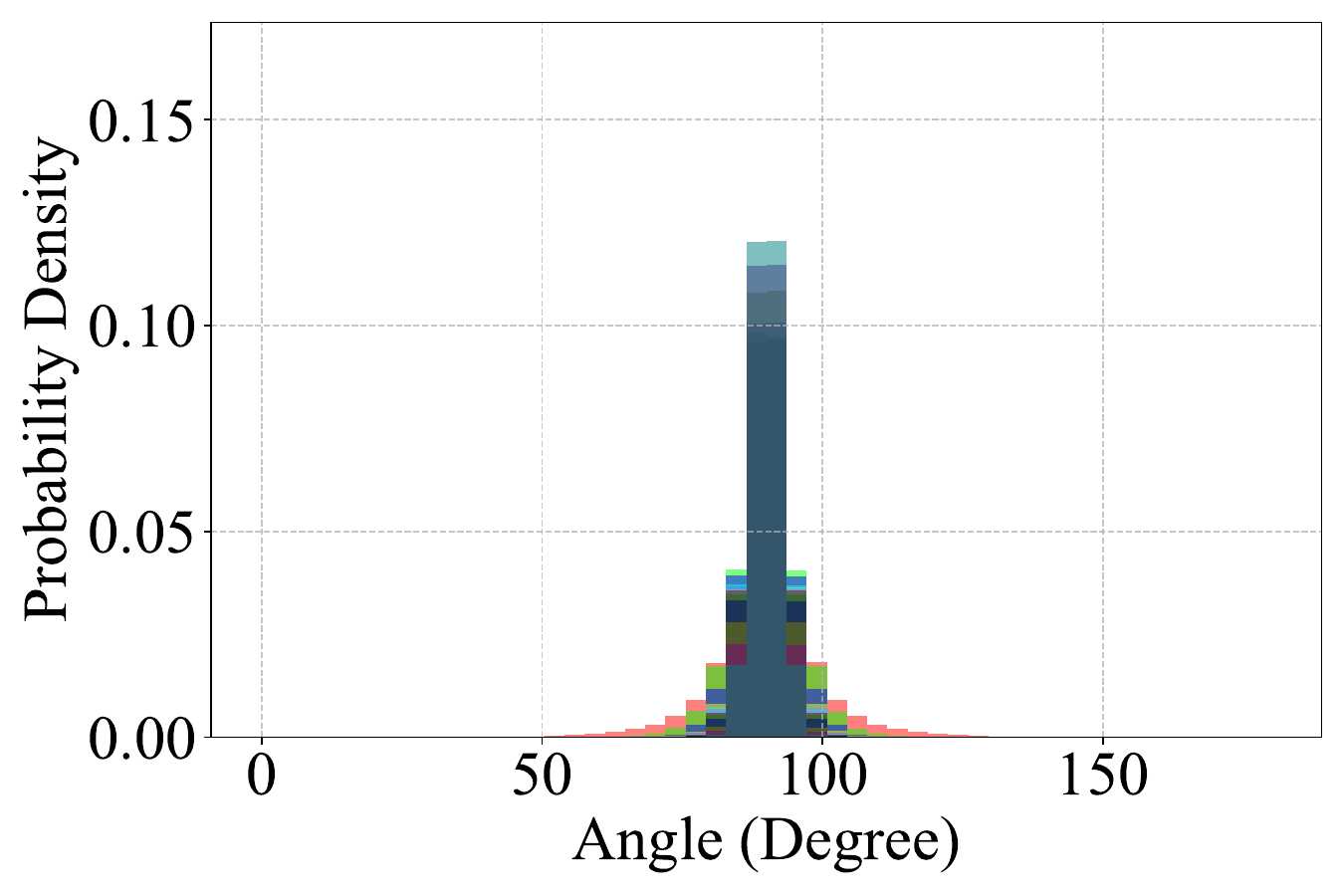} 
        \caption{The angle distribution of pre-trained matrices $\mathbf{W}_{q}$.}
        \label{fig:vit-query-pre-train}
    \end{subfigure}
    \begin{subfigure}[b]{0.22\textwidth}
        \includegraphics[width=\textwidth]{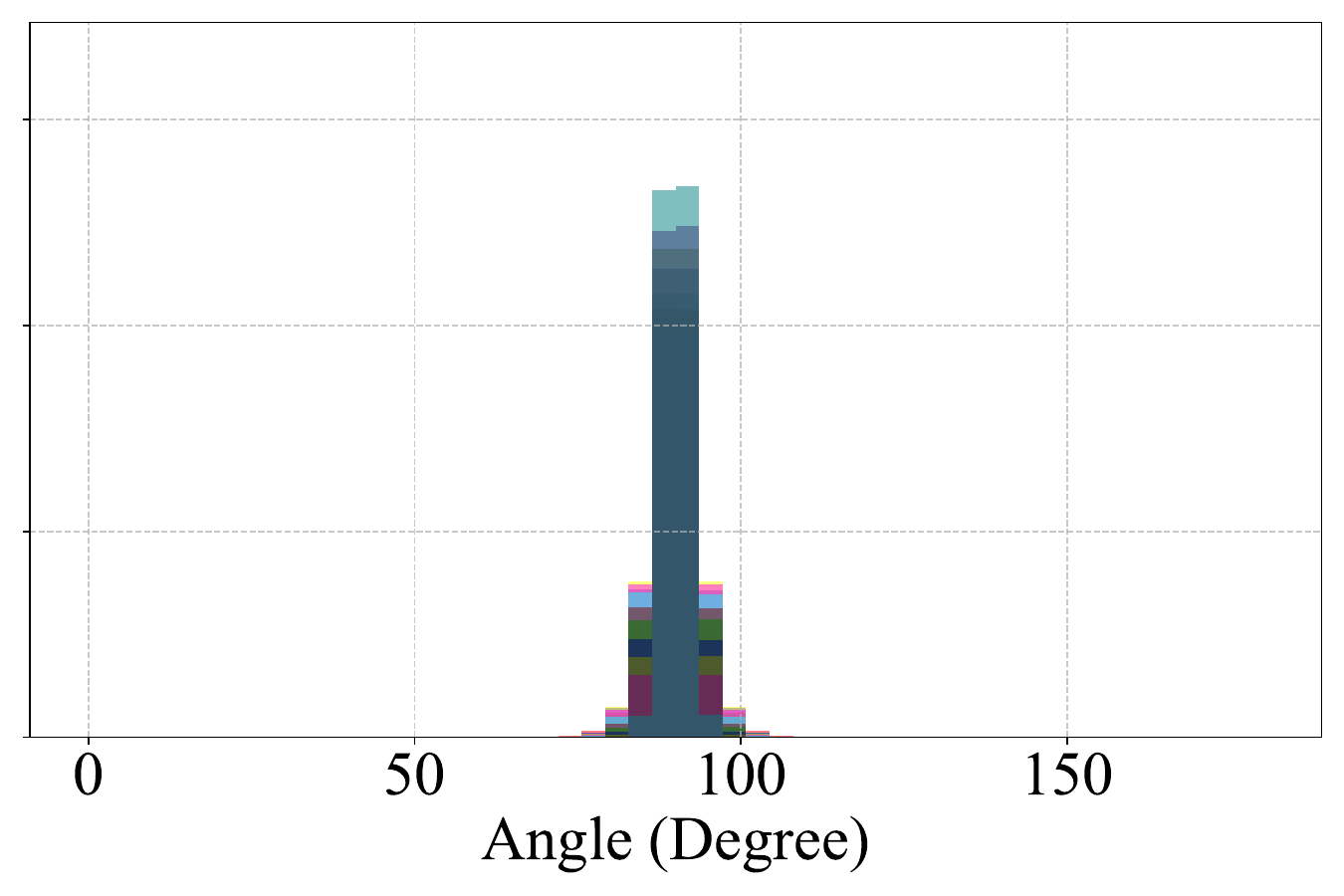} 
        \caption{The angle distribution of pre-trained matrices $\mathbf{W}_{v}$.}
        \label{fig:vit-value-pre-train}
    \end{subfigure}
    \vspace{-0.3cm}
    \caption{Illustration of approximate orthogonality among any two column vectors of weight matrices $\mathbf{W}_{q}, \mathbf{W}_{v},$ in the ViT-B model. The histogram represents the distribution of angles between any two column vectors within each weight matrix. Specifically, 
    \subref{fig:vit-query-pre-train} and \subref{fig:vit-value-pre-train} represent approximate orthogonality in the pre-trained model. The whole results and their legends can be found in Appendix~\ref{sec:Legend}.}
    \label{fig:vit-init-pre-train}
    \vspace{-0.7cm}
\end{figure}

Upon closer examination of the backbone parameters of ViTs, an interesting observation emerges: the weight matrices within the backbone parameters after pre-training exhibit an approximate orthogonality among any two row or column vectors, as shown in Fig.~\ref{fig:vit-init-pre-train}. Orthogonality is a fundamental property in linear algebra that signifies the independence of vectors within a matrix. When any two vectors in the weight matrix are approximately orthogonal, it subtly suggests that the model has a smaller upper bound on generalization error and thus possesses an enhanced generalization capability. The detailed analysis of the upper bound on generalization error is shown in Section~\ref{sec:upper-bound}.

\begin{figure}[!tb]
    \centering
    \begin{subfigure}[b]{0.22\textwidth}
        \includegraphics[width=\textwidth]{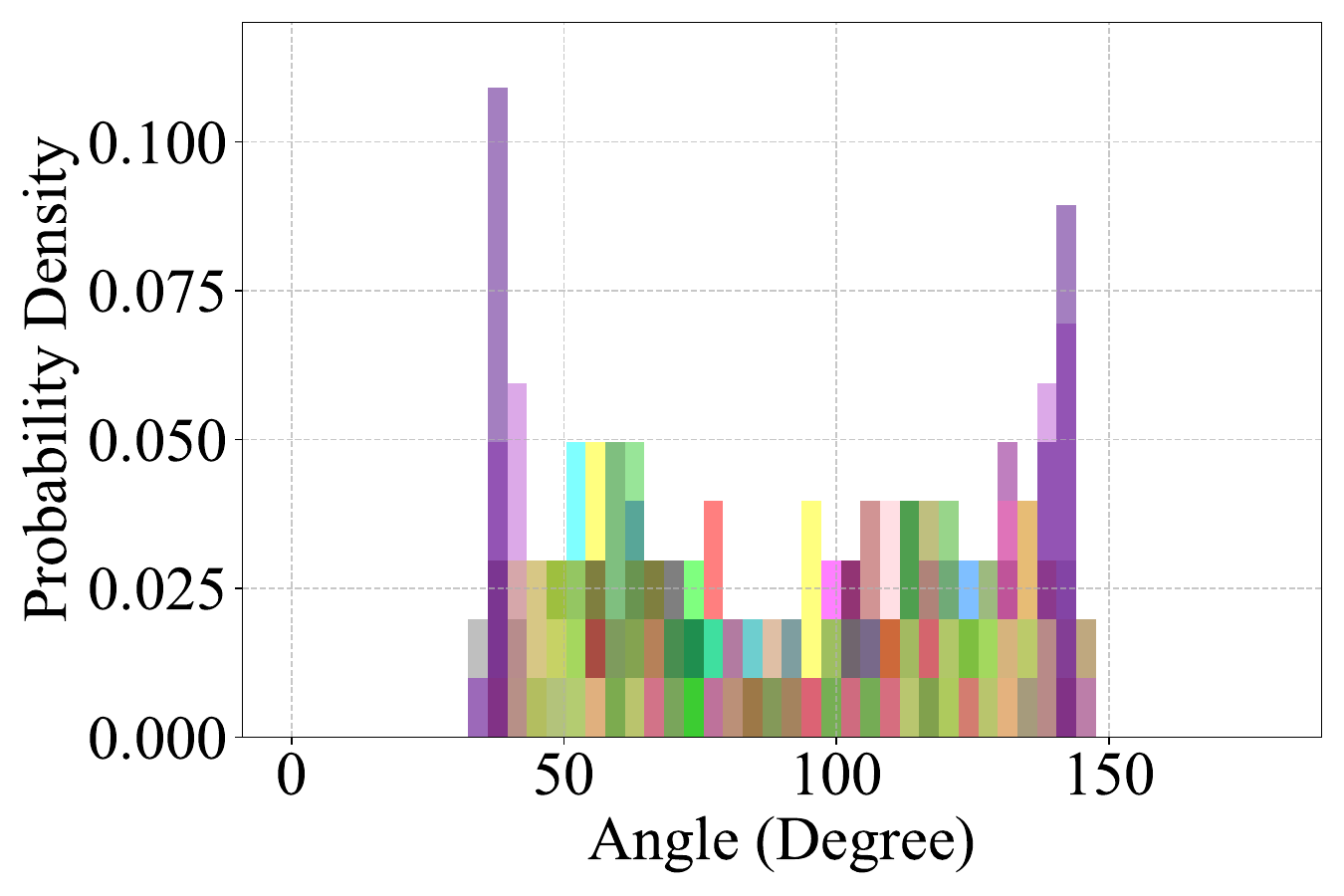} 
        \caption{The angle distribution of any two column vectors of down-projection matrix in LoRA.}
        \label{fig:lora-a}
    \end{subfigure}
    \begin{subfigure}[b]{0.22\textwidth}
        \includegraphics[width=\textwidth]{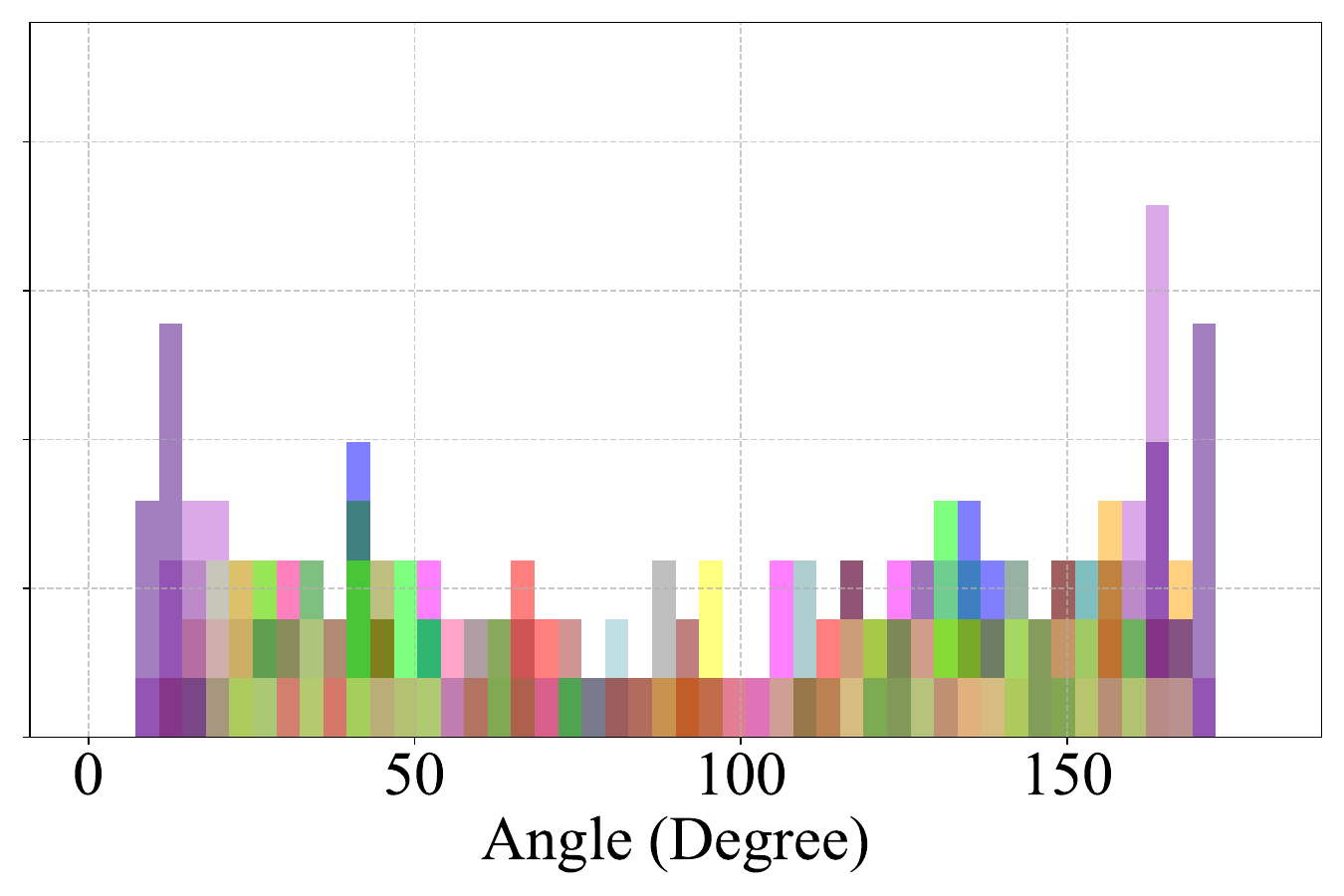} 
        \caption{The angle distribution of any two column vectors of up-projection matrix in LoRA.}
        \label{fig:lora-b}
    \end{subfigure}
    \begin{subfigure}[b]{0.22\textwidth}
        \includegraphics[width=\textwidth]{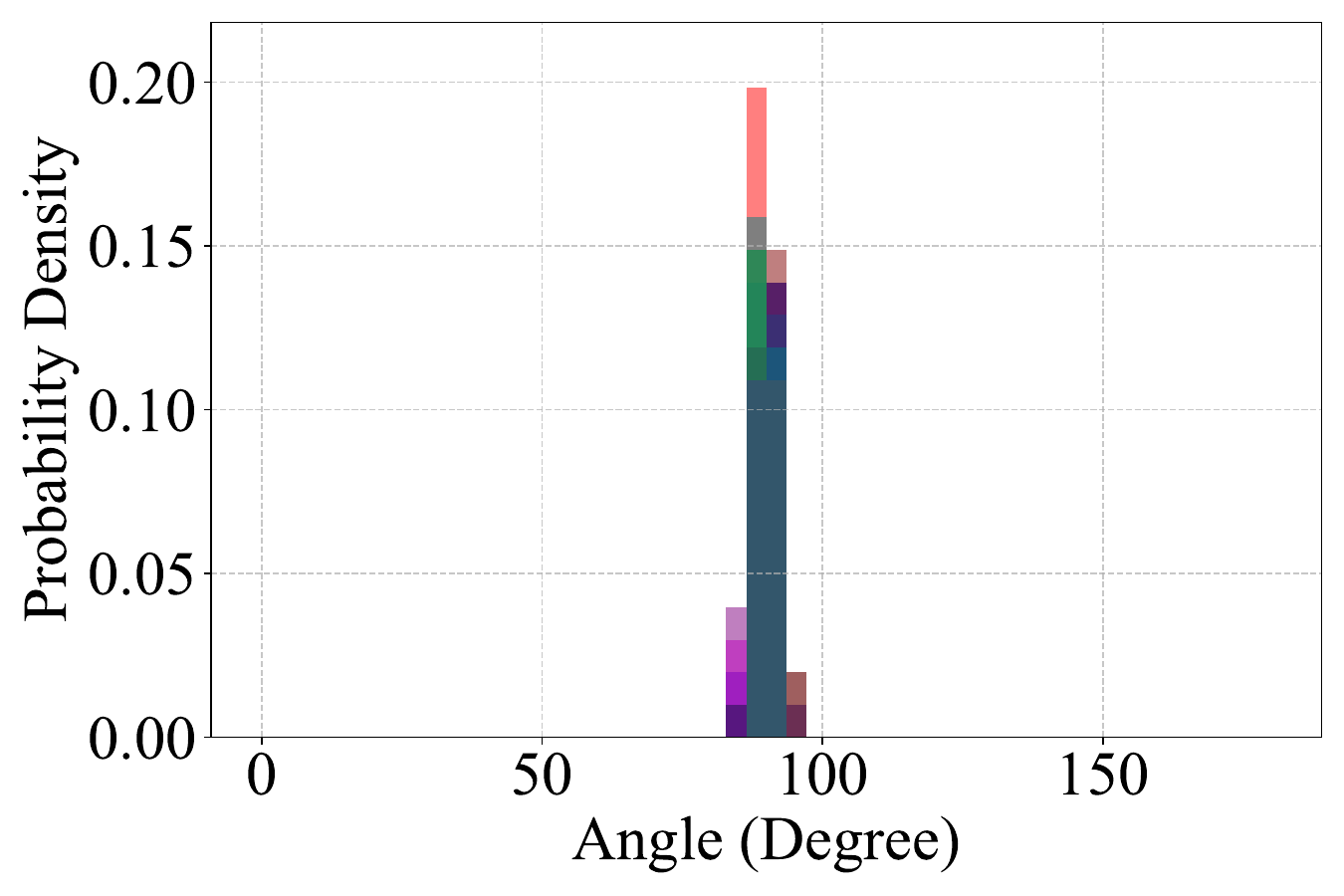} 
        \caption{The angle distribution of any two column vectors of down-projection matrix in Adapter.}
        \label{fig:adapter-down}
    \end{subfigure}
    \begin{subfigure}[b]{0.22\textwidth}
        \includegraphics[width=\textwidth]{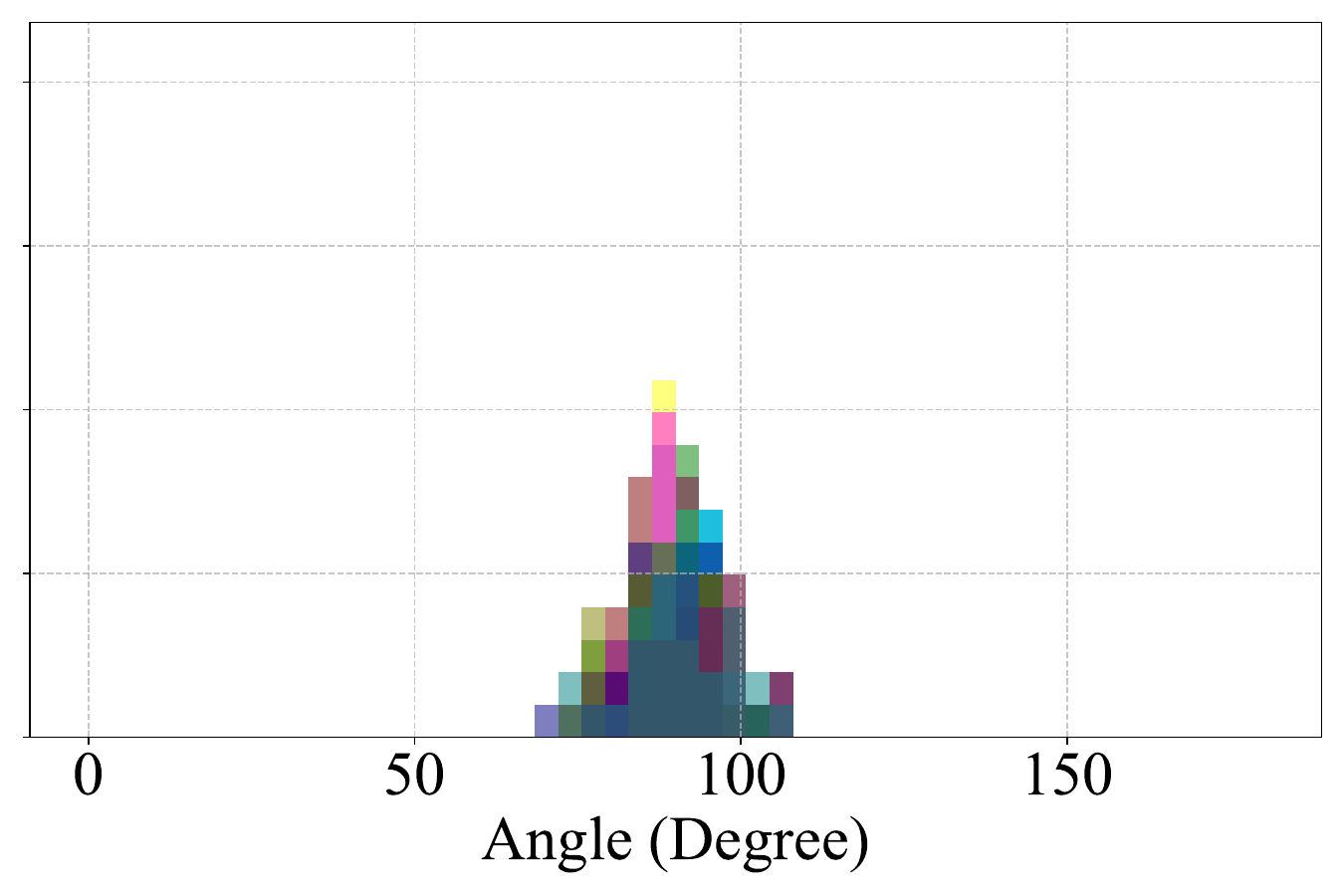} 
        \caption{The angle distribution of any two column vectors of up-projection matrix in Adapter.}
        \label{fig:adapter-up}
    \end{subfigure}
    \vspace{-0.3cm}
    \caption{The angle distribution between any two column vectors within the down/up-projection matrices in the LoRA and Adapter structures, where \subref{fig:lora-a} and \subref{fig:lora-b} represent the results trained using LoRA method, \subref{fig:adapter-down} and \subref{fig:adapter-up} represent the results trained using Adapter method. These results are all obtained from models trained on the dtd dataset. Their legends can be found in Appendix~\ref{sec:Legend}.}
    \label{fig:lora-adapter}
    \vspace{-0.7cm}
\end{figure}

However, this desirable orthogonal property is notably absent in the vectors of the down/up-projection matrices used in current PEFT~\cite{hu2021lora, houlsby2019parameter} methods, as shown in Fig.~\ref{fig:lora-adapter}. This discrepancy raises an intriguing question: if the down/up-projection matrices were to exhibit the same approximate orthogonality as the backbone matrices, could the generalization capability of fine-tuned ViTs be further augmented? In other words, could aligning the properties of these adaptation matrices with those of the backbone lead to improved performance in downstream tasks?

To explore this hypothesis, we propose a novel Approximately Orthogonal Fine-Tuning (AOFT) strategy for representing the low-rank weight matrices in PEFT. Our approach aims to align the properties of the down/up-projection matrices with those of the backbone matrices by enforcing approximate orthogonality. Specifically, we employ a single learnable vector to generate a set of approximately orthogonal vectors, which are then used to construct the down/up-projection matrices. The rationale behind this strategy is grounded in the theoretical understanding of orthogonal matrices. Orthogonal matrices are known for their numerical stability, efficient computation, and ability to preserve the norm of vectors~\cite{golub2013matrix, trefethen2022numerical}. When approximate orthogonal weight matrices are used to replace the down/up-projection matrices, the upper bound of the model's generalization error decreases, thereby enhancing the model's generalization capability.

To validate the efficacy of our AOFT strategy, we conducted extensive experiments across a range of downstream image classification tasks. Our results demonstrate that the proposed method achieves competitive performance compared to existing PEFT techniques. These findings not only confirm the hypothesis that enforcing orthogonality in the down/up-projection matrices can augment the generalization capability of fine-tuned ViTs but also highlight the potential of our AOFT strategy as a powerful tool for PEFT in computer vision. In summary, the contributions of this work can be summarized as follows:

\begin{itemize}
    \item We introduce a novel AOFT strategy for aligning the properties of the down/up-projection matrices with those of the backbone matrices by enforcing approximate orthogonality, leveraging the theoretical understanding of orthogonal matrices and their beneficial properties for model representation.
    \item We utilized a learnable vector to generate down/up-projection matrices with approximately orthogonal row/column vectors, thereby reducing the number of learnable parameters and consequently decreasing the overhead of fine-tuning the model.
    \item Through extensive experiments across a range of downstream image classification tasks, we demonstrates that the proposed AOFT strategy achieves competitive performance compared to existing PEFT techniques.
\end{itemize}

\section{Related Work}
\label{sec:formatting}

\subsection{Parameter-Effcient Fine-Tuning}
In contrast to the full fine-tuning technique, which involves learning all parameters, parameter-efficient fine-tuning modifies only a small subset of parameters while keeping the majority of parameters in the model's backbone frozen. This methodology enables the model to efficiently adapt to downstream tasks, while concurrently achieving substantial reductions in training costs. In the field of PEFT methods, numerous works continue to emerge, constantly driving technological innovation and development~\cite{kopiczko2023vera}. Among them, Adapter method~\cite{houlsby2019parameter}, as one of the mainstream techniques for fine-tuning large models, introduces an innovative fine-tuning paradigm by inserting trainable adapter components into the network structure to achieve efficient model adjustment. The Low-Rank Adaptation (LoRA)~\cite{hu2021lora} method designs the adapter as a side path and cleverly utilizes a bottleneck structure to significantly reduce the number of learnable parameters during fine-tuning. This design not only alleviates the computational burden on the model but also simulates incremental changes in the parameter matrix. Visual Prompt Tuning (VPT)~\cite{jia2022visual} adds a small number of prompt parameters to the input layer and intermediate layers of ViT, and only trains these prompt parameters while keeping the backbone frozen. Adapter Re-Composing (ARC)~\cite{dong2023efficient} employs a unique approach using symmetric up-and-down projections to create cross-layer shared bottleneck operations. By learning low-dimensional rescaling coefficients, it effectively reconfigures layer-adaptive adapters, thereby reducing the cost of fine-tuning. Residual-based Low-Rank Rescaling (RLRR)~\cite{dong2024low} analyzes PEFT methods from the perspective of Singular Value Decomposition (SVD), ensuring through a residual design that new parameters do not deviate excessively from the pre-trained model.

These methods do not introduce approximate orthogonality into their weight matrices to preserve the model's generalization capability.

\subsection{PEFT via Orthogonality}
Orthogonal matrices have certain applications in fine-tuning strategies. Orthogonal Fine-Tuning (OFT)~\cite{qiu2023controlling} preserves the pretrained semantics and concepts by mapping the linear weights with an angle-preserving transformation using a block-diagonal matrix, while adapting to downstream tasks. Quasi-Givens Orthogonal Fine-Tuning (qGOFT)~\cite{ma2024parameter} adjusts the pre-trained weight matrix to accommodate different types of downstream tasks by left-multiplying it with a Givens Rotation matrix. Orthogonal Butterfly (BOFT)~\cite{liu2023parameter} parameterizes dense orthogonal matrices using the product of multiple sparse orthogonal matrices based on butterfly decomposition. 

Our method generates an entire approximately orthogonal down/up-projection  matrices through a single vector. This approach is simple and efficient, discarding the fine-tuning of down/up-projection matrices using approximately orthogonal matrices employed by existing methods, thereby enhancing the model's generalization capability while improving efficiency.


\section{Methodology}
In this section, we first outline the concepts related to the PEFT methods. Then, we propose a method to replace projection matrices in LoRA and Adapter methods with down/up-projection matrices in which column/row vectors are approximately orthogonal; all these column/row vectors are generated by one vector. Finally, we analyze the effectiveness of approximately orthogonal projection matrices in preserving the model's generalization capability.

\begin{figure*}[!tb]
    \centering
    \includegraphics[width=0.95\linewidth]{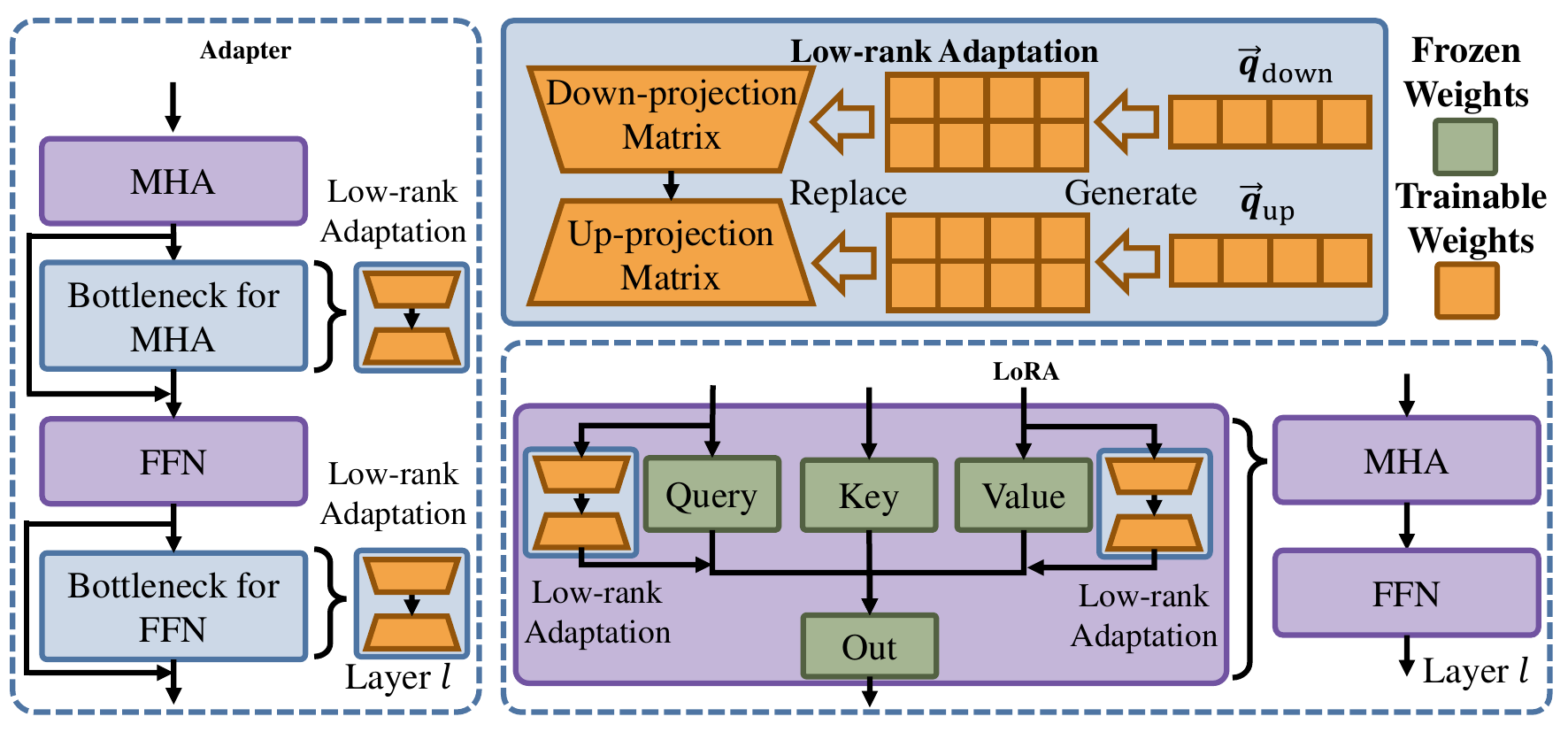}
    \caption{Illustration of the proposed AOFT method: We construct an approximately orthogonal matrix using a learnable vector to replace the up-projection and down-projection matrices in the bottleneck structure. To validate its effectiveness, we apply it to the LoRA and Adapter architectures. (1) Within the Adapter architecture, which itself incorporates a Low-Rank Adaptation structure, we employ a residual approach to fine-tune the frozen pre-trained parameter matrices for any weight matrix in the Multi-Head Attention (MHA) and Feed-Forward Network (FFN) modules. (2) Similarly, in the LoRA framework, we utilize dynamically generated down-projection and up-projection matrices instead of their counterparts applied to the $\mathbf{W}_q$ and $\mathbf{W}_v$. }
    \label{fig:Method}
    \vspace{-0.5cm}
\end{figure*}


\subsection{Preliminary Knowledge}
Vision Transformer (ViT)~\cite{dosovitskiy2020image} is a deep learning model that applies the Transformer architecture to computer vision tasks such as image classification. ViT consists of two main components: the Patch Embedding layer and the Transformer Encoder layer. The Patch Embedding layer splits an input image $\mathbf{X} \in \mathbb{R}^{H \times W \times C}$ into fixed-size patches and projects each patch into a sequence;
the Transformer Encoder processes the patch embeddings using Multi-Head Attention (MHA) and Feed-Forward Network (FFN) blocks. The computation of the $(l)$-th Transformer Encoder layer is defined as:
\begin{equation}
    \begin{split}
\mathbf{X}^{(l)'} &= {\rm MHA}({\rm LN}(\mathbf{X}^{(l-1)})) + \mathbf{X}^{(l-1)}, \\
\mathbf{X}^{(l)} &= {\rm FFN}({\rm LN}(\mathbf{X}^{(l)'})) + \mathbf{X}^{(l)'},
\end{split}
\end{equation}
where $\rm{LN}(\cdot)$ is a function to layer representation normalization. In the MHA block, the computation of each attention head is defined as:
\begin{equation}
    \begin{split} 
    &\mathrm {AttentionHead}(\mathbf{X}^{(l-1)}) = \\
    &{\rm Softmax}\left(\frac{(\mathbf{X}^{(l-1)} \mathbf{W}_q^{(l)}) (\mathbf{X}^{(l-1)} \mathbf{W}_k^{(l)})^\top}{\sqrt{D_\mathrm{head}^{(l)}}}\right) \mathbf{X}^{(l-1)} \mathbf{W}_v^{(l)},
    \end{split}
\end{equation}
where $\mathbf{W}_q^{(l)}, \mathbf{W}_k^{(l)}, \mathbf{W}_v^{(l)} \in \mathbb{R}^{D^{(l-1)} \times D_\mathrm{head}^{(l)}}$ are the query, key, and value weight matrices, respectively, $D_\mathrm{head}^{(l)}$ is the output dimension of each attention head. MHA is the weighted sum of multiple attention heads according to the weights $\mathbf{W}_o^{(l)}$.

The output of the MHA block is normalized and fed into the FFN block:
\begin{equation}
	{\rm FFN}(\mathbf{X}^{(l)})={\rm GELU}(\mathbf{X}^{(l)}\mathbf{W}_{\mathrm {FC1}}^{(l)})\mathbf{W}_{\mathrm {FC2}}^{(l)},
\end{equation}
where $\mathbf{W}_{\mathrm {FC1}}^{(l)} \in \mathbb{R}^{D^{(l)} \times 4 \cdot D^{(l)}}$ and $\mathbf{W}_{\mathrm {FC2}}^{(l)} \in \mathbb{R}^{4 \cdot D^{(l)} \times D^{(l)}}$ are two linear projection matrices.

\subsection{Approximately Orthogonal Fine-Tuning}
In~\cite{mayer2013simple} a method to construct the orthogonal matrix is proposed, and the detailed construction method is provided in~\cref{sec:operator_Q}. Following this construction, we use a learnable vector $\vec{\boldsymbol q}=(q_0, q_1, \cdots, q_N)^\top \in \mathbb{R}^{N}$ to generate an orthogonal matrix $\mathbf{Q}\in \mathbb{R}^{N\times N}$, then, we replace down- or up-projection matrix with the orthogonal matrix $\mathbf{Q}$. This generation operator is defined as:
\begin{equation}
    \mathrm{AO}(\vec{\boldsymbol q})=\mathbf{Q}[:, 0:d],
\end{equation}
\noindent where the $d$ is the dimension of the bottleneck in low-rank adaptation and $[:,:]$ represents indexes for the matrix, used to obtain a submatrix from that matrix. The orthogonal matrix $\mathbf{Q}$ is defined as:
\begin{equation}
\begin{bmatrix}
    q_0 & -q_1 & -q_2 & \cdots & -q_N \\
    q_1 & 1-\frac{q_1q_1}{1+q_0} & -\frac{q_2q_1}{1+q_0} & \cdots & -\frac{q_Nq_1}{1+q_0} \\
    q_2 & -\frac{q_1q_2}{1+q_0} & 1-\frac{q_2q_2}{1+q_0} & \cdots & -\frac{q_Nq_2}{1+q_0} \\
    \vdots & \vdots & \vdots & \vdots & \vdots \\
    q_i & -\frac{q_1q_i}{1+q_0} & -\frac{q_2q_i}{1+q_0} & \cdots & -\frac{q_Nq_i}{1+q_0} \\
    \vdots & \vdots & \vdots & \vdots & \vdots \\
    q_N & -\frac{q_1q_N}{1+q_0} & -\frac{q_2q_N}{1+q_0} & \cdots & 1-\frac{q_Nq_N}{1+q_0}
\end{bmatrix},
\label{Eq:AOFT}
\end{equation}
\noindent the column vectors of $\mathbf{Q}$ are strictly orthogonal to each other when using the normalization $\sum_{i=1}^N|q_i|^2=1$. To enhance the flexibility of the model's capacity, we do not strictly adhere to this normalization, allowing the column vectors of the matrix $\mathbf{Q}$ to be approximately orthogonal. 

\begin{figure}[!tb]
    \centering
    \begin{subfigure}[t]{0.22\textwidth}
        \includegraphics[width=\textwidth]{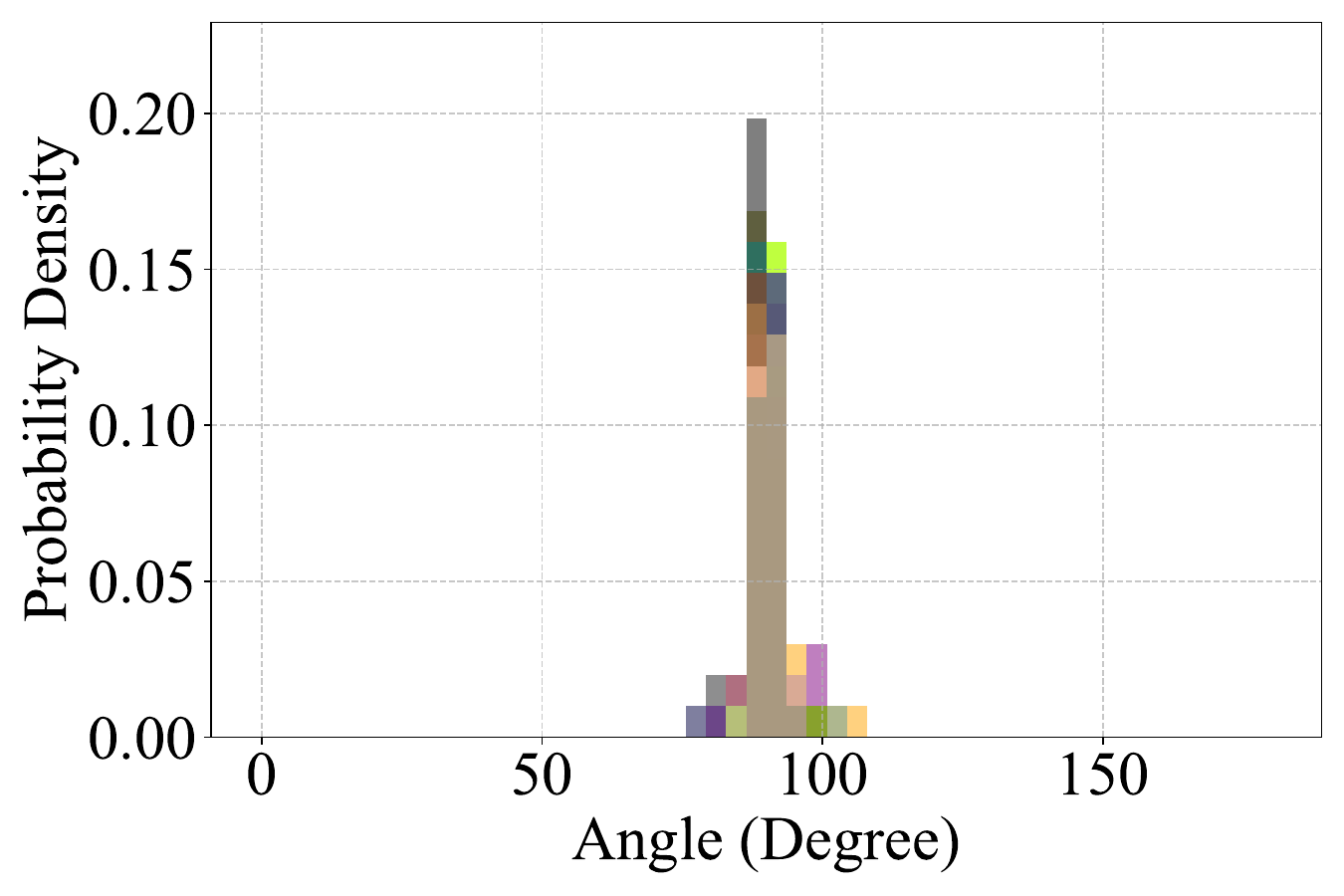} 
        \caption{The angle distribution of any two column vectors of down-projection matrix in LoRA with AOFT.}
        \label{fig:lora-down-ours}
    \end{subfigure}
    \begin{subfigure}[t]{0.22\textwidth}
        \includegraphics[width=\textwidth]{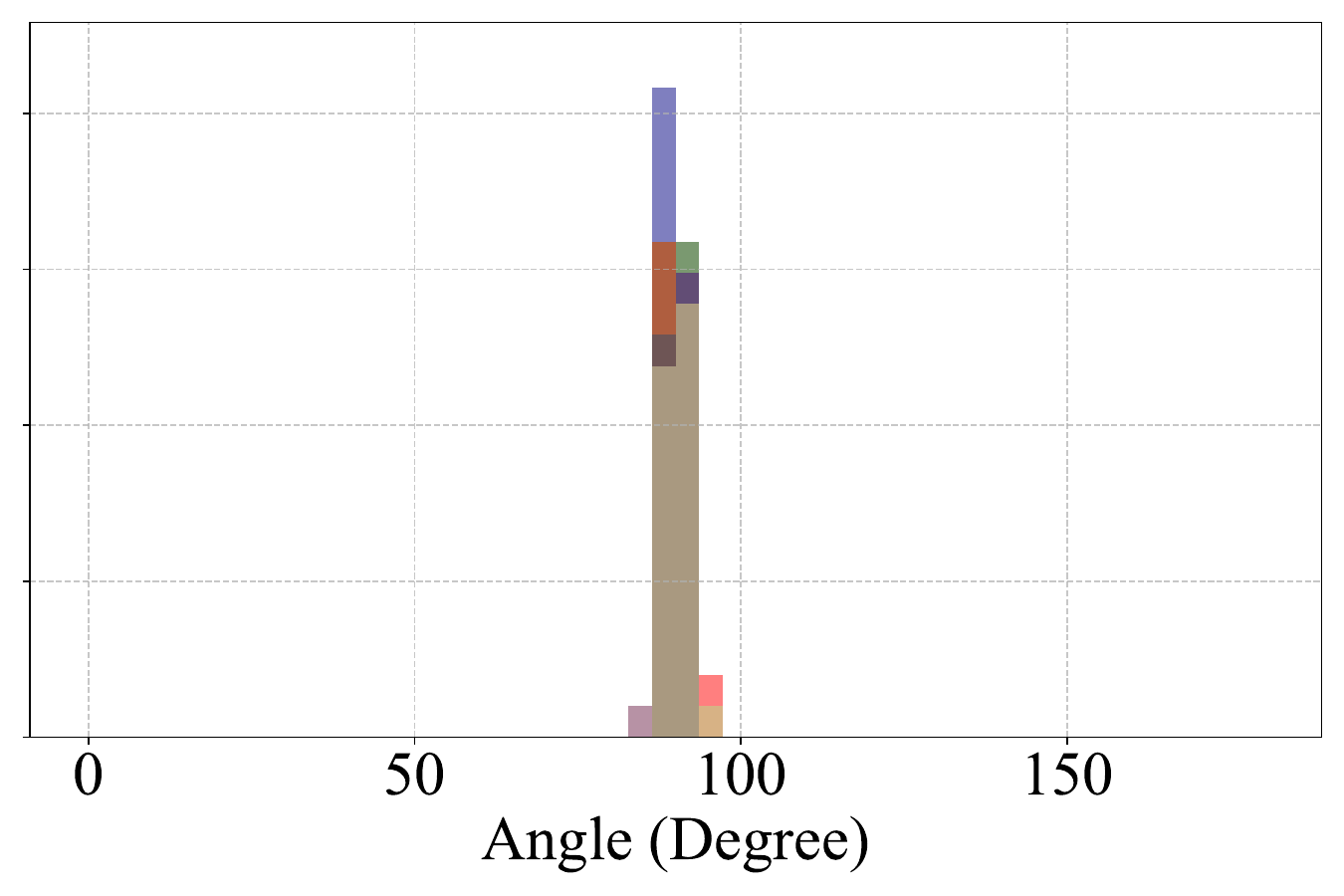} 
        \caption{The angle distribution of any two column vectors of up-projection matrix in LoRA with AOFT.}
        \label{fig:lora-up-ours}
    \end{subfigure}
    \begin{subfigure}[t]{0.22\textwidth}
        \includegraphics[width=\textwidth]{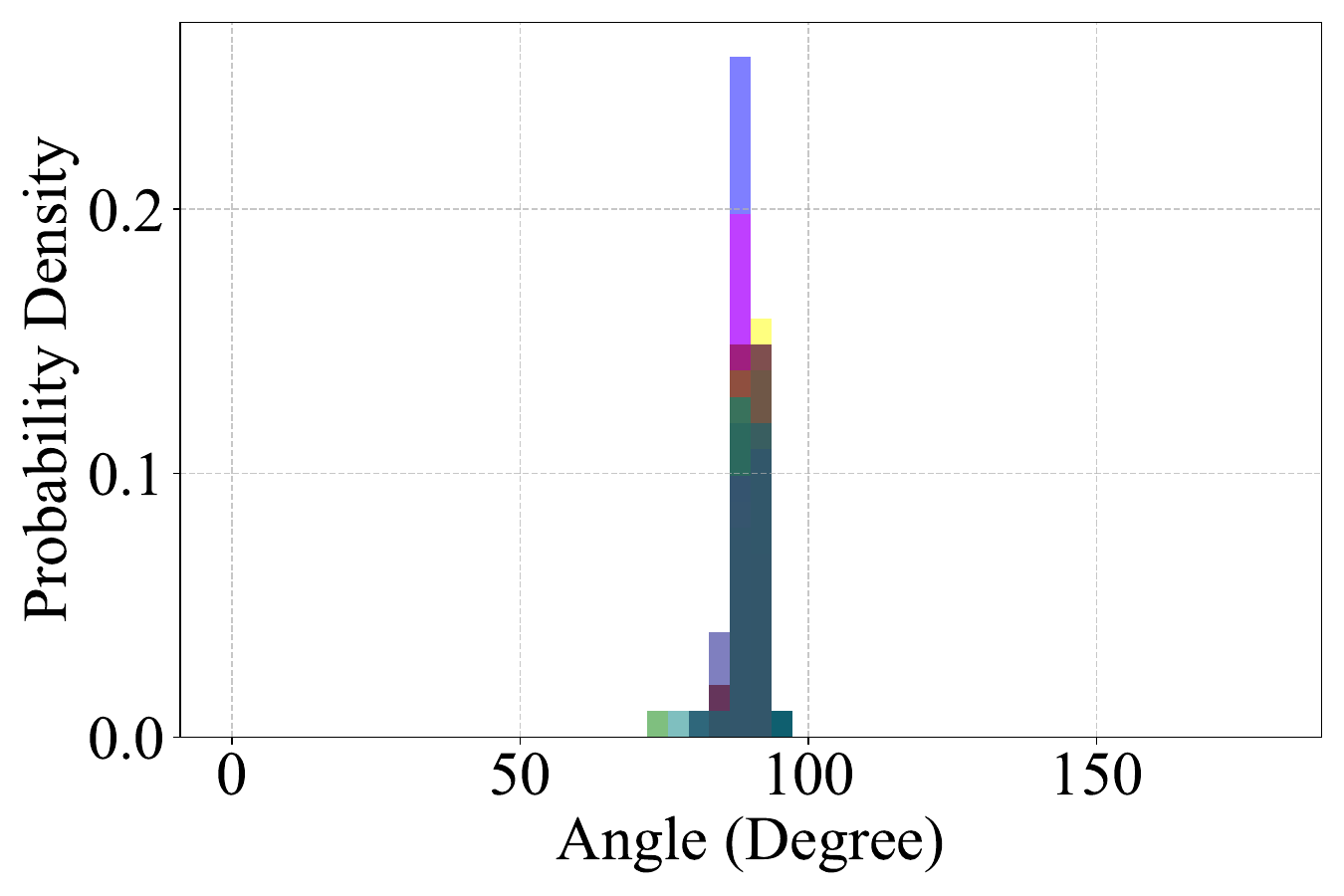} 
        \caption{The angle distribution of any two column vectors of down-projection matrix in Adapter with AOFT.}
        \label{fig:adapter-down-ours}
    \end{subfigure}
    \begin{subfigure}[t]{0.22\textwidth}
        \includegraphics[width=\textwidth]{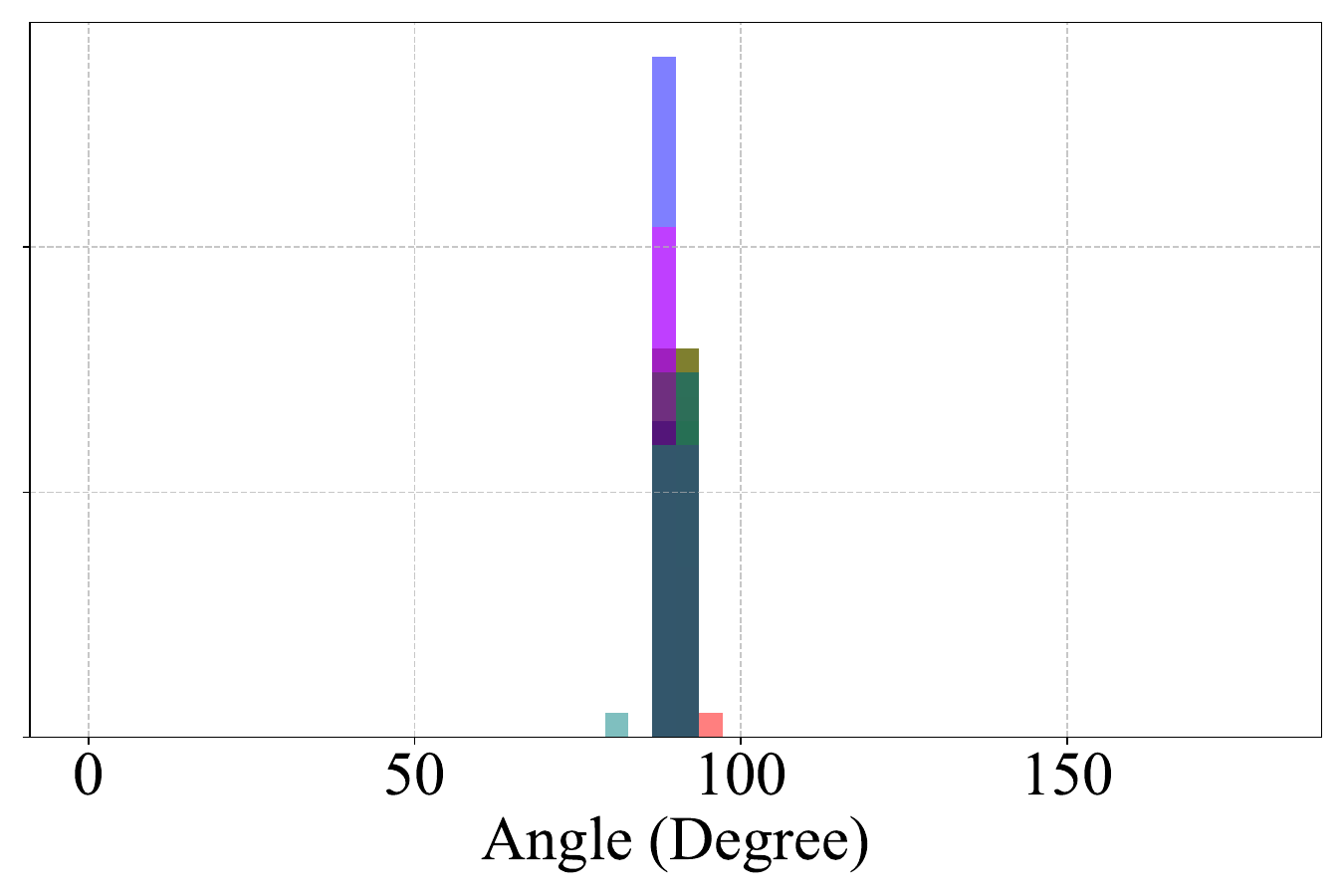} 
        \caption{The angle distribution of any two column vectors of up-projection matrix in Adapter with AOFT.}
        \label{fig:adapter-up-ours}
    \end{subfigure}
    \vspace{-0.2cm}
    \caption{Down/up-projection matrices trained through using the combination of AOFT with LoRA, Adapter methods exhibit a distribution of angles between pairs of vectors that are concentrated around 90 degrees, indicating that these vectors are approximatively pairwise orthogonal to each other. Specifically, \subref{fig:lora-down-ours} and \subref{fig:lora-up-ours} show the angle distribution of any two column vectors in fine-tuned down/up-projection matrices of LoRA; \subref{fig:adapter-down-ours} and \subref{fig:adapter-up-ours} demonstrate the counterparts of Adapter. 
    These figures are all obtained from models trained on the dtd dataset.Their legends can be found in Appendix~\ref{sec:Legend}.}
    \label{fig:lora-adapter-ours}
    \vspace{-0.5cm}
\end{figure}

Based on this design, we can derive the LoRA alternative as follows:
\begin{equation}
    \begin{split} 
    \mathbf{X}_{\mathrm{FT}}^{(l-1)}=\mathbf{X}^{(l-1)}(\mathbf{W}^{(l)}+\mathrm {AO}(\vec{\boldsymbol q}_{\mathrm{down}}){\mathrm {AO}(\vec{\boldsymbol q}_\mathrm{up})}^\top),
    \end{split}
    \label{eq:AO_LoRA}
\end{equation}
\noindent where $\mathbf{W}^{(l)}$ is $\mathbf{W}_{q}^{(l)}$ or $\mathbf{W}_{v}^{(l)}$. Similarly, we can derive a scheme for combining AOFT with Adapter-based method, as follows:
\begin{equation}
    \begin{split} 
    \mathbf{X}_{\mathrm{FT}}^{(l-1)}=&\mathrm{MHA}(\mathbf{X}^{(l-1)})\mathrm {AO}(\vec{\boldsymbol q}_{\mathrm{down}}^\mathrm{MHA}) \mathrm {AO}{(\vec{\boldsymbol q}_\mathrm{up}^\mathrm{MHA})}^\top, \\
    \mathbf{X}_{\mathrm{FT}}^{(l)}=&\mathrm{FFN}(\mathbf{X}_{\mathrm{FT}}^{(l-1)})\mathrm {AO}(\vec{\boldsymbol q}_{\mathrm{down}}^\mathrm{FFN}) \mathrm {AO}{(\vec{\boldsymbol q}_\mathrm{up}^\mathrm{FFN})}^\top. \\
    \end{split}
    \label{eq:AO_adapter}
\end{equation}
\noindent Their AOFT structures are shown in Fig.~\ref{fig:Method}, and we can observe that the degree distribution of any two column vectors within the fine-tuned projection matrix reveals approximate orthogonality, as shown in \cref{fig:lora-adapter-ours}. 

Despite the differences between the prompt structure of VPT and the bottleneck structure of LoRA and Adapter, we also try to apply AOFT to the VPT approach. We replace the prompt part with a single matrix, as follows:
\begin{equation}
    \begin{aligned}
     \mathbf{X}_{\mathrm{FT}}^{(l-1)}= &\left(\begin{array}{l}
	\mathbf{X}^{(l-1)} \\
    \mathrm{AO}(\vec{\boldsymbol q}_\mathrm{prompt})^{\top(l-1)} \end{array}\right)\mathbf{W}^{(l)}+\vec{\boldsymbol{b}}^{(l)\top},
    \end{aligned}
    \label{eq:AO_prompt}
\end{equation}
\noindent and the angle distribution of any two column vectors of prompt matrix is illustrated in Fig.~\ref{fig:vpt-ours} of Section~\ref{sec:ablation}.

\subsection{Upper Bound of Generalization Error for AOFT}\label{sec:upper-bound}

Generalization error reflects the generalization capability of a model. If one model has a smaller generalization error than another, then the model is considered effective. In fact, the generalization error represents the expected risk of the learned model.

In this section, we utilize Rademacher Complexity~\cite{goar2024foundations} to define the upper bound of generalization error of models:
\vspace{-0.2cm}
\begin{equation}
\begin{split}
    &\mathbb{E}_{\xi\in\{\pm 1\}^m}\left[\frac{1}{m}{\rm sup}_{\lVert \mathbf{W} \rVert\leq \gamma}\lVert \sum_{i=1}^{m}\xi_i\mathbf{W}\vec{\boldsymbol x}_{i} \rVert\right] \\
    &= \mathbb{E}_{\xi\in\{\pm 1\}^m}\left[\frac{1}{m}{\rm sup}_{\lVert \mathbf{W} \rVert\leq \gamma}\lVert \mathbf{W}\sum_{i=1}^{m}\xi_i\vec{\boldsymbol x}_{i} \rVert\right] \\
    &\leq \mathbb{E}_{\xi\in\{\pm 1\}^m}\left[\frac{1}{m}{\rm sup}_{\lVert \mathbf{W} \rVert\leq \gamma}\lVert \mathbf{W}\rVert\lVert\sum_{i=1}^{m}\xi_i\vec{\boldsymbol x}_{i} \rVert\right], \\
\label{eq:RC_L2}
\end{split}
\end{equation}
\vspace{-0.3cm}

\noindent where $m$ represents the number of samples; $\vec{\boldsymbol x}_{i}$ denotes the sample features; $\mathbf{W}$ is the weight matrix; $\xi_i$ is a Rademacher variable; and $\gamma$ stands for the L2-norm of $\mathbf{W}$. Hence, the magnitude of $\gamma$ determines the upper bound of the generalization error. As $\gamma$ increases, the upper bound of the generalization error becomes larger; conversely, as $\gamma$ decreases, the upper bound of the generalization error becomes smaller.

\begin{figure}[!tb]
    \centering
    \begin{subfigure}[t]{0.235\textwidth}
        \includegraphics[width=\textwidth]{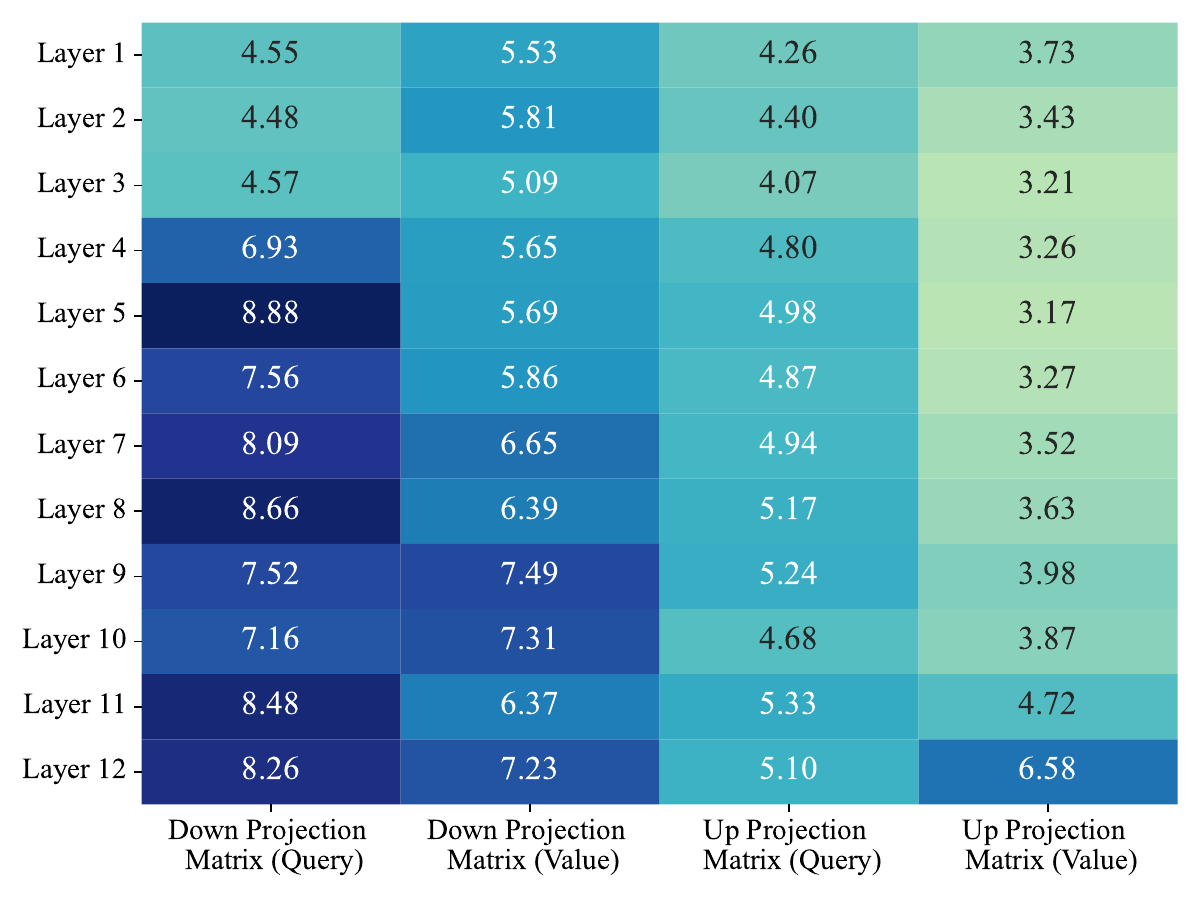} 
        \caption{The L2-norms of down/up-projection matrices in LoRA.}
        \label{fig:heatmap-lora-base}
    \end{subfigure}
    \begin{subfigure}[t]{0.235\textwidth}
        \includegraphics[width=\textwidth]{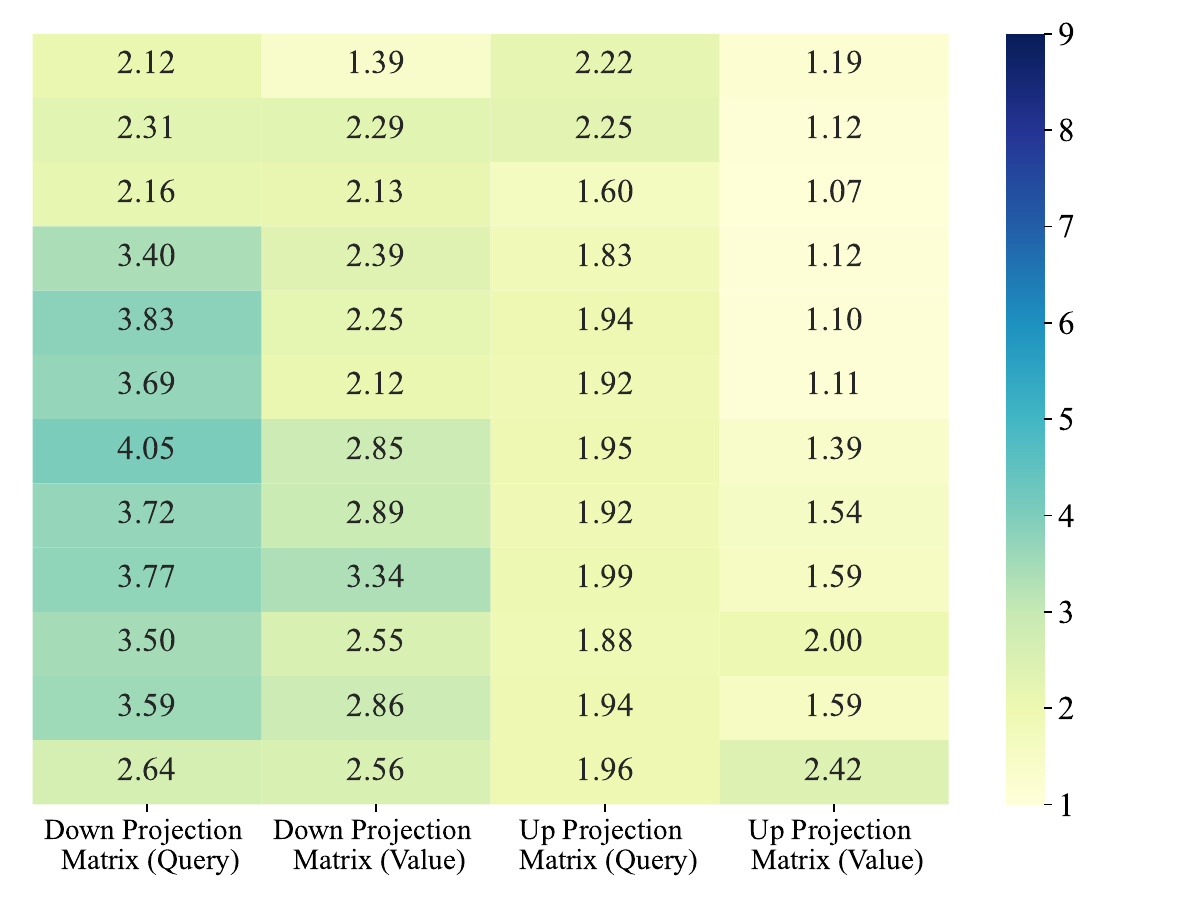} 
        \caption{The L2-norms of down/up-projection matrices in LoRA with AOFT.}
        \label{fig:heatmap-lora-ours}
    \end{subfigure}
    \begin{subfigure}[t]{0.235\textwidth}
        \includegraphics[width=\textwidth]{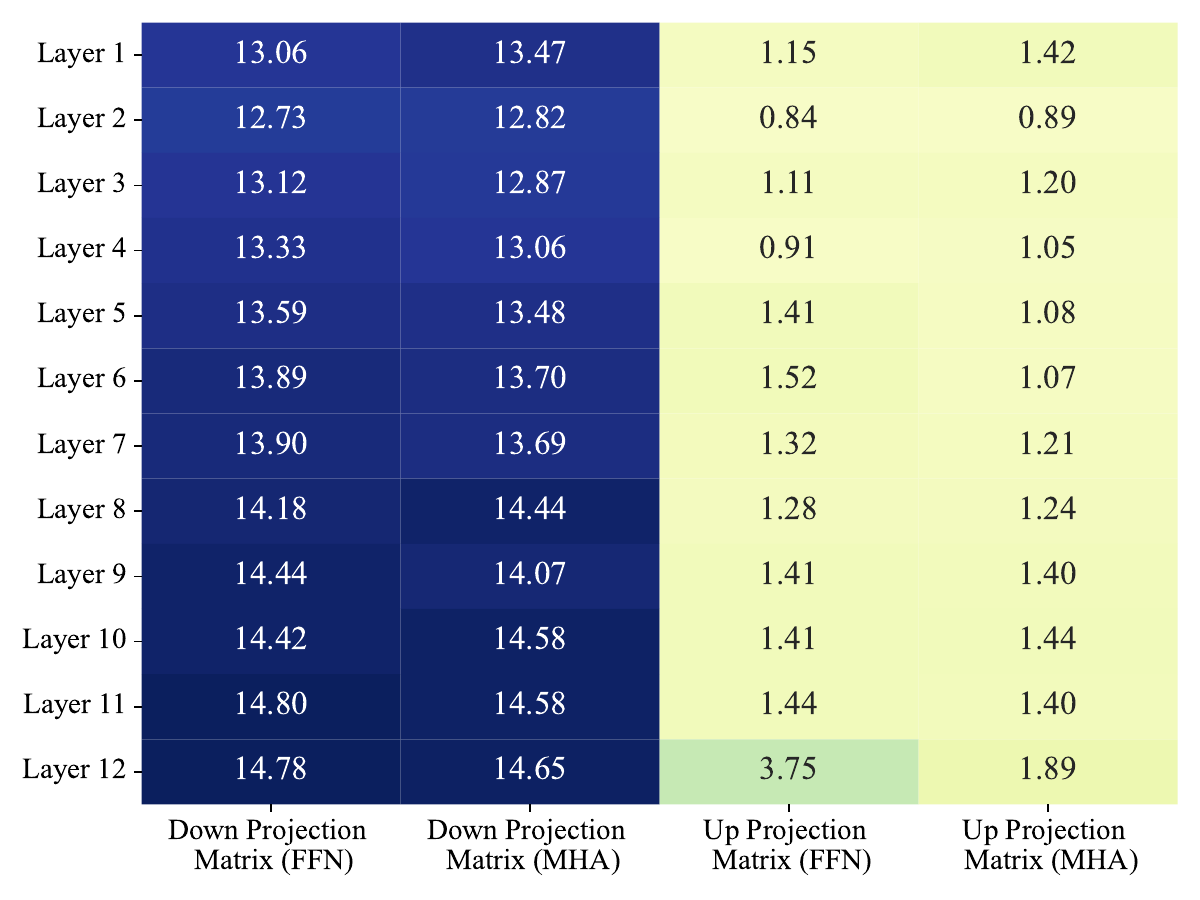} 
        \caption{The L2-norm of down/up-projection matrices in Adapter.}
        \label{fig:heatmap-lora-ours}
    \end{subfigure}
    \begin{subfigure}[t]{0.235\textwidth}
        \includegraphics[width=\textwidth]{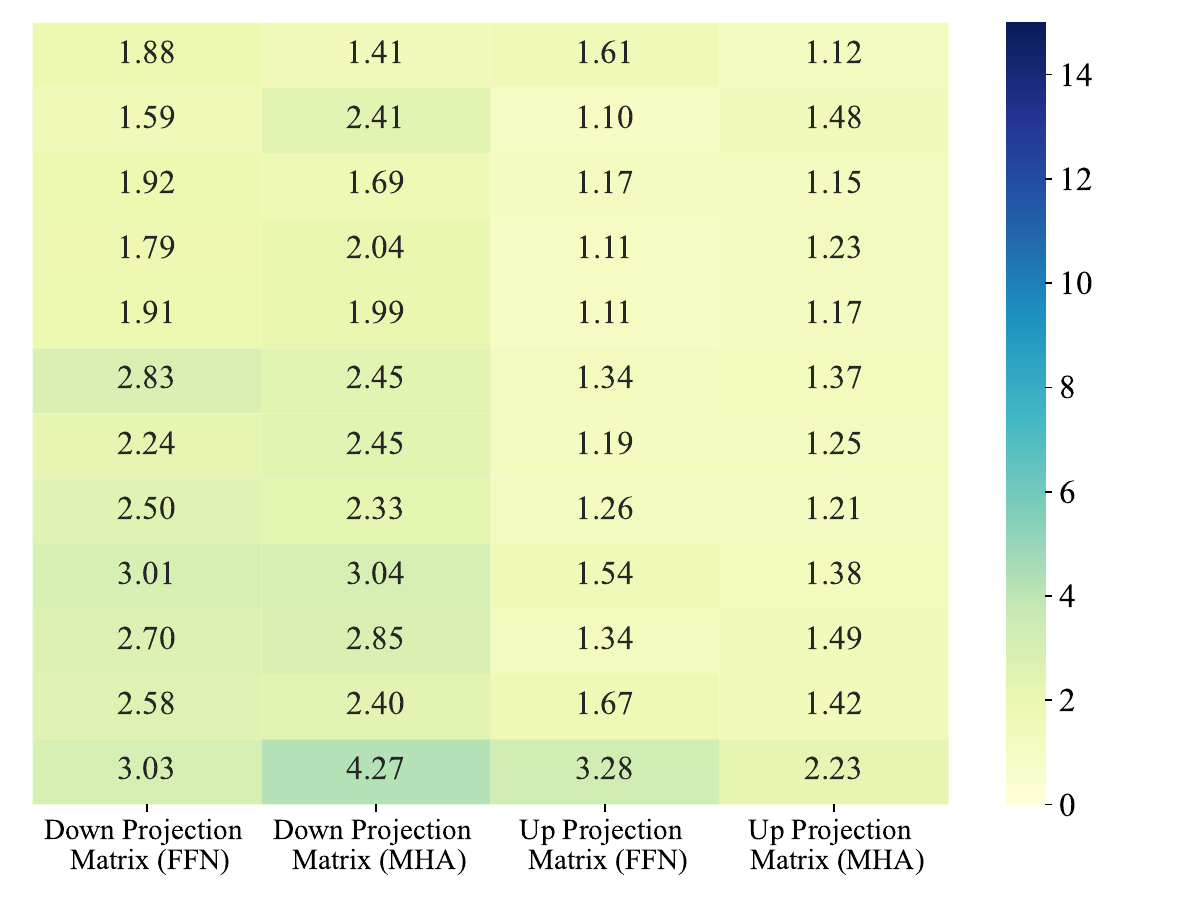} 
        \caption{The L2-norm of down/up-projection matrices in Adapter with AOFT.}
        \label{fig:heatmap-lora-ours}
    \end{subfigure}
    \vspace{-0.4cm}
    \caption{Comparison of L2-norms between the LoRA/Adapter method and the LoRA/Adapter method enhanced with AOFT.}
    \label{fig:heatmap}
    \vspace{-0.4cm}
\end{figure}

In our method, the L2-norms $\lVert \mathbf{W}_{\rm down} \rVert$ and $\lVert \mathbf{W}_{\rm up} \rVert$ of AOFT are significantly smaller than that of the LoRA and Adapter methods, as illustrated in Fig.~\ref{fig:heatmap}. Consequently, the upper bound of the generalization error for AOFT is much lower than that of the Lora and Adapter methods, demonstrating that AOFT possesses superior generalization capability.

\section{Experiments}
\label{sec:exp}
We conducted a comprehensive series of experiments, demonstrating that AOFT is a robust and reliable method with plug-and-play functionality. The detailed experimental setups and additional results are presented in the supplementary materials.
\begin{table*}[!tb] \scriptsize
	\centering
        \caption{Performance comparison of Adapter, LoRA and VPT, where AOFT is integrated into these PEFT methods on the FGVC dataset. All experiments are conducted using ViT-B/16 pre-trained on ImageNet-21K as the backbone. The best results are shown in \textbf{bold}.}
        \vspace{-0.2cm}
        \resizebox{16.5cm}{!}{
		\begin{tabular}{c|c|c|c|c|c|cc}
			\toprule
			\diagbox{\textbf{Methods}}{\textbf{Datasets}} & \multicolumn{1}{c|}{\textbf{CUB-200-2011}} & \textbf{NABirds} & \multicolumn{1}{c|}{\textbf{Oxford Flowers}} & \multicolumn{1}{c|}{\textbf{Stanford Dogs}} & \multicolumn{1}{c|}{\textbf{Stanford Cars}} & \textbf{Mean Total} & \multicolumn{1}{c}{\textbf{Params. (M)}} \\
			\midrule
			Full fine-tuning & 87.3  & 82.7  & 98.8  & 89.4  & 84.5  & 88.5  &  85.98 \\
			Linear probing & 85.3  & 75.9  & 97.9  & 86.2  & 51.3  & 79.3  &  0.18 \\
                \midrule
			Adapter~\cite{houlsby2019parameter} & 87.1  & 84.3  & 98.5  & 89.8  & 68.6  & 85.7  &  0.41 \\
			Adapter+AOFT  & 88.6  & 83.7  & 99.4  & 89.8  & 83.0  & 88.9  &  0.20 \\
                Adapter+AOFT*  & \textbf{89.0}  & \textbf{84.5}  & \textbf{99.5}  & \textbf{92.0}  & \textbf{85.2}  & \textbf{90.1}  & 0.20  \\
                \midrule
			LoRA~\cite{hu2021lora}  & 88.3  & \textbf{85.6} & 99.2  & 91.0  & 83.2  & 89.5  &  0.44 \\
                LoRA+AOFT   & \textbf{88.8} & 84.2 & 99.4 & \textbf{92.0} & \textbf{85.1} & \textbf{89.9} &  0.22 \\
			LoRA+AOFT*   & 88.4 & 84.2 & \textbf{99.5} & \textbf{92.0} & 85.0 & 89.8 &  0.22 \\
                \midrule
                VPT-Shallow~\cite{jia2022visual} & 86.7  & 78.8  & 98.4  & 90.7  & \textbf{68.7}  & \textbf{84.7}  & 0.25  \\
                VPT-Shallow+AOFT(dim=64) & \textbf{88.9}  & \textbf{81.5} & \textbf{99.8} &  \textbf{92.4} & 61.7  & 84.0  & 0.18  \\
                \midrule
                VPT-Deep~\cite{jia2022visual} & 88.5  & \textbf{84.2}  & 99.0  & 90.2  & 83.6  & 89.1  & 0.85  \\
                VPT-Deep+AOFT(dim=768) & \textbf{88.7} & 82.8 & \textbf{99.5} & \textbf{91.5} & \textbf{84.1} & \textbf{89.5} & 0.15 \\
			\bottomrule
		\end{tabular}%
    }
	\label{vitb_fgvc}%
    \vspace{-0.4cm}
\end{table*}%

\subsection{Experiments Setting}

\textbf{Datasets.} Following the previous works~\cite{jia2022visual, lian2022scaling, dong2023efficient, jie2023fact}, we utilized two visual adaptation benchmarks to assess the effectiveness of our method: Fine-Grained Visual Classification (FGVC) and VTAB-1k~\cite{zhai2019large}, encompassing a total of 24 datasets. FGVC comprises five Fine-Grained Visual Classification datasets: \textit{CUB-200-2011}~\cite{wah2011caltech}, \textit{NABirds}~\cite{van2015building}, \textit{Oxford Flowers}~\cite{nilsback2008automated}, \textit{Stanford Dogs}~\cite{khosla2011novel}, and \textit{Stanford Cars}~\cite{gebru2017fine}. Meanwhile, VTAB-1k includes a diverse set of datasets across various categories. These datasets are specifically tailored to tackle the intricate task of distinguishing between visually similar subcategories within broader categories, thereby rendering the challenge more demanding and intricate. The VTAB-1k benchmark showcases an impressive array of 19 diverse visual classification tasks, categorized into three broad groups: \textit{Natural}, \textit{Specialized}, and \textit{Structured}. \textit{Natural} focuses on images captured by standard cameras, representing natural scenes. \textit{Specialized} features images taken by specialized equipment, such as remote sensing and medical imaging devices. \textit{Structured} includes synthesized images from simulated environments, covering tasks like object counting and 3D depth prediction. Each VTAB-1k task comprises 1,000 training samples. The relevant content can be found in the Appendix~\ref{sec:rationale}.


\noindent
\textbf{Baselines and existing PEFT methods.} To evaluate the performance of AOFT, we conducted a comparative analysis by integrating it with two baseline methods and three well-known PEFT approaches including LoRA, Adapter, and VPT-Shallow/Deep. The two baseline methods are: (1) Full Fine-Tuning, which involves updating all parameters of the pre-trained model using training data from the downstream task, and (2) Linear Probing, which entails training only a linear classification head on the downstream task while keeping the remaining pre-trained parameters frozen.

\noindent
\textbf{Our PEFT methods.} The AOFT methods are denoted as Adapter+AOFT (Eq.~\ref{eq:AO_adapter}), LoRA+AOFT (Eq.~\ref{eq:AO_LoRA}), VPT-Shallow+AOFT (Eq.~\ref{eq:AO_prompt}) and VPT-Deep+AOFT (Eq.~\ref{eq:AO_prompt}). To enhance the flexibility of our method, we introduce a learnable scaling vector~$\vec{\boldsymbol \lambda}$ to modulate the down/up-projection matrix, forming AOFT* for both Adapter and LoRA. Specifically, AOFT* is defined as $(\mathbf{W}_{\rm down}\odot\vec{\boldsymbol \lambda}^\top)\mathbf{W}_{\rm up}$, where $\odot$ denotes element-wise multiplication. Since our approach does not introduce additional parameters when increasing the bottleneck dimension, we can dynamically adjust the bottleneck size for different downstream tasks, ensuring optimal performance for each setting. The corresponding results are reported in the AOFT$^{\dag}$ row of Tab.~\ref{vitb_vtab}.

\noindent
\textbf{Implementation details.} During the training phase, we employed the default data augmentation strategy. For the FGVC datasets, images were randomly resized and cropped to $224 \times 224$ pixels, and a random horizontal flip was applied to augment the data. For the VTAB-1k datasets, we adhered to the default settings and directly resized the images to $224 \times 224$ pixels. To fine-tune the models, we employed the AdamW optimizer for 100 epochs. Additionally, we managed the learning rate using a cosine decay strategy, ensuring effective model training. All experiments are conducted using the PyTorch framework on an NVIDIA A800 GPU with 80 GB of memory.

\begin{table*}[!tb]
	\centering
        \caption{Performance comparison of Adapter, LoRA, VPT, OFT and GOFT where AOFT is integrated into these PEFT methods on the VTAB-1k dataset. All experiments are conducted using ViT-B/16 pre-trained on ImageNet-21K as the backbone. The best results are shown in \textbf{bold}. AOFT* is defined as $(\mathbf{W}_{\rm down}\odot\vec{\boldsymbol \lambda}^\top)\mathbf{W}_{\rm up}$,  AOFT$^{\dag}$ indicates results using different bottleneck structures.}
        \vspace{-0.2cm}
	  \resizebox{\linewidth}{!}{
		\begin{tabular}{c|ccccccc|c|cccc|c|cccccccc|c|cc}
			\toprule
			\multirow{2}[2]{*}{\diagbox{\textbf{Methods}}{\textbf{\rotatebox{0}{Datasets}}}} & \multicolumn{8}{c|}{\textbf{Natural}}                         & \multicolumn{5}{c|}{\textbf{Specialized}} & \multicolumn{9}{c|}{\textbf{Structed}}                                &       &  \\
			& \rotatebox{90}{\textbf{CIFAR-100}} & \rotatebox{90}{\textbf{Caltech101}} & \rotatebox{90}{\textbf{DTD}} & \rotatebox{90}{\textbf{Flowers102}} & \rotatebox{90}{\textbf{Pets}} & \rotatebox{90}{\textbf{SVNH}} & \multicolumn{1}{c}{\rotatebox{90}{\textbf{Sun397}}} & \rotatebox{90}{\textbf{Mean}} & \rotatebox{90}{\textbf{Camelyon}} & \rotatebox{90}{\textbf{EuroSAT}} & \rotatebox{90}{\textbf{Resisc45}} & \multicolumn{1}{c}{\rotatebox{90}{\textbf{Retinopathy}}} & \rotatebox{90}{\textbf{Mean}} & \rotatebox{90}{\textbf{Clevr-Count}} & \rotatebox{90}{\textbf{Clevr-Dist}} & \rotatebox{90}{\textbf{DMLab}} & \rotatebox{90}{\textbf{KITTI-Dist}} & \rotatebox{90}{\textbf{dSpr-Loc}} & \rotatebox{90}{\textbf{dSpr-Ori}} & \rotatebox{90}{\textbf{sNORB-Azim}} & \multicolumn{1}{c}{\rotatebox{90}{\textbf{sNORB-Ele}}} & \rotatebox{90}{\textbf{Mean}} & \rotatebox{90}{\textbf{Mean Total}} & \rotatebox{90}{\textbf{Params.(M)}} \\
			\midrule
			Full fine-tuning & 68.9  & 87.7  & 64.3  & 97.2  & 86.9  & 87.4  & 38.8  & 75.9  & 79.7  & 95.7  & 84.2  & 73.9  & 83.4  & 56.3  & 58.6  & 41.7  & 65.5  & 57.5  & 46.7  & 25.7  & 29.1  & 47.6  & 65.6  & 85.80  \\
			Linear probing & 63.4  & 85.0  & 63.2  & 97.0  & 86.3  & 36.6  & 51.0  & 68.9  & 78.5  & 87.5  & 68.6  & 74.0  & 77.2  & 34.3  & 30.6  & 33.2  & 55.4  & 12.5  & 20.0  & 9.6   & 19.2  & 26.9  & 52.9  & 0.04  \\
		  \midrule
			VPT-Shallow~\cite{jia2022visual} & \textbf{77.7 } & 86.9  & 62.6  & 97.5  & 87.3  & 74.5  &  51.2  &  76.8  &  78.2  &  92.0  &  75.6  &  72.9  &  79.7  &  50.0  &  58.6  &  40.5  &  67.1  &  \textbf{68.7}  &  36.1  &  20.2  &  \textbf{34.1}  &  47.0  &  64.8  &  0.11 \\
            VPT-Shallow+AOFT(dim=64)                 & 71.8  &  \textbf{90.7} & \textbf{71.8}  & \textbf{99.0} & \textbf{91.3}  & \textbf{75.6} &  \textbf{56.7}  & \textbf{79.5}   &  \textbf{81.5}  &  \textbf{92.6}  &  \textbf{81.3}  &  \textbf{74.3}  &  \textbf{82.4}  &  \textbf{53.0}  &  \textbf{59.0}  &  \textbf{44.2}  &  \textbf{76.4}  &  58.6  &  \textbf{40.7}  &  \textbf{22.5}  &  28.1  & \textbf{47.8}  &  \textbf{66.8}  & 0.05  \\
                \midrule
                VPT-Deep~\cite{jia2022visual} & \textbf{78.8} & 90.8  & 65.8  & 98.0  & 88.3  & 78.1  & 49.6  & 78.5  & 81.8  & \textbf{96.1}  & 83.4  & 68.4  & 82.4  & \textbf{68.5}  & 60.0  & 46.5  & 72.8  & \textbf{73.6}  & 47.9  & \textbf{32.9}  & \textbf{37.8}  & 55.0  & 69.4  & 0.60  \\
                VPT-Deep+AOFT(dim=768) & 70.7 & \textbf{92.9}  & \textbf{70.1} & \textbf{99.2} & \textbf{91.5} &  \textbf{87.0}  &  \textbf{50.8}   &  \textbf{80.3}  & \textbf{84.9} &  93.9   &  \textbf{83.9}  &  \textbf{76.1}  &  \textbf{84.7}  &  65.6  &  \textbf{61.2}  &  \textbf{48.6}  & \textbf{82.3}   & 72.1   & \textbf{49.2}   & 27.9   &  36.0  & \textbf{55.4}  &  \textbf{70.7}  &  0.05 \\
                 \midrule
            Adapter~\cite{houlsby2019parameter} & 69.2  & 90.1  & 68.0  & 98.8  & 89.9  & 82.8  &  54.3  & 79.0   &  \textbf{84.0}  &  \textbf{94.9}  &  81.9  &  75.5  &  84.1  &  \textbf{80.9}  &  \textbf{65.3}  &  \textbf{48.6}  &  78.3  &  74.8  &  48.5  &  \textbf{29.9}  &  41.6  & 58.5   &  71.4  &  0.16 \\
			Adapter+AOFT  & 64.9  & 93.3  & \textbf{71.6}  & 99.1  & \textbf{90.9}  & \textbf{85.5}  &  49.7  &  79.3  &  83.4  &  94.2  &  83.3  &  \textbf{76.0}  &  \textbf{84.2}  &  79.4  &  63.2  &  47.6  &  \textbf{81.3}  &  \textbf{87.0}  &  \textbf{54.7}  &  26.3  &  \textbf{45.4}  &  \textbf{60.6}  & 72.5  &  0.06  \\
                Adapter+AOFT*  & \textbf{74.6}  & \textbf{93.9}  &  71.4 & \textbf{99.4}  & 90.1  & 78.0  &  \textbf{57.1}  &  \textbf{81.4}  &  82.0  &  94.0  &  \textbf{84.1}  &  75.6  &  83.9  &  78.8  &  62.5  &  48.5  &  79.6  &  84.6  &  54.3  &  24.1  &  42.7  & 59.4   & \textbf{72.7}  &  0.06  \\
                \midrule

			LoRA~\cite{hu2021lora}  & 67.1  & 91.4  & 69.4  & 98.8  & 90.4  & \textbf{85.3}  &  54.0  & 79.5   &  \textbf{84.9}  &  \textbf{95.3}  &  \textbf{84.4}  &  73.6  & 84.6   &  \textbf{82.9}  &  \textbf{69.2}  &  \textbf{49.8}  &  78.5  &  75.7  &  47.1  &  \textbf{31.0}  &  44.0  &  \textbf{59.8}  &  72.3  &  0.29 \\
                LoRA+AOFT   & 73.6  & \textbf{92.6}  & 71.1  & \textbf{99.3}  & \textbf{91.3}  & \textbf{85.3}  &  \textbf{56.9}  & \textbf{81.4}   &  84.7  &  94.8  &  83.7  &  \textbf{75.6}  &  \textbf{84.7}  &  76.7  &  63.2  &  48.7  &  81.0  &  82.2  &  53.1  &  26.9  &  \textbf{45.1}  & 59.6   &  \textbf{72.9}  &  0.08 \\
			LoRA+AOFT*   & \textbf{74.0}  & 91.0  & \textbf{72.7}  & \textbf{99.3}  & 89.3  & 80.6  &  56.8  &  80.5  &  \textbf{84.9}  &  94.6  &  82.7  &  \textbf{75.6}  &  84.4  &  71.4  &  57.5  &  42.7  &  \textbf{82.0}  &  \textbf{83.4}  &  \textbf{53.9}  &  22.6  &  44.5  &  57.3  &  71.5  &  0.08 \\
             \midrule
             VeRA~\cite{kopiczko2023vera} &  \textbf{74.5} & 92.7  &	71.5  & 99.2 & 	90.0 & 	84.8 & 	56.2 & 	81.3 & 	82.5 & 	92.0  &	 81.5 & 	74.7 & 	82.7 & 	71.5 & 	61.3 & 	49.3 & 	79.5 & 	78.7 & 	42.9 & 	30.5 & 	42.4 & 	57.0 & 	71.4 &  0.03 \\
             OFT~\cite{qiu2023controlling} & 65.1 &	88.4  &	70.0  &	96.5  &	88.7  &	\textbf{88.5}  &	45.7  &	77.6  &	\textbf{86.7}  &	\textbf{96.6}  &	\textbf{85.5}  &	77.0  &	\textbf{86.5} &	79.7  &	67.6  &	50.2 &	77.8  &	78.5  &	50.8  &	30.9  &	40.5  &	59.5  &	71.8  &	0.15 \\
             DoRA~\cite{liu2024dora} & 66.1 & 93.1 & 68.8 & 97.0 & 89.9 & 87.2& 56.4 & 79.8 & 83.1& 94.5& 80.8& 75.2 & 83.4 & 79.1& 62.1 & 48.0 & 80.6  & 83.1& 51.8& \textbf{33.2}& 44.0 & 60.2 & 72.3 & 0.19 \\
             GOFT~\cite{ma2024parameter} & 71.5 &	90.9  &	68.8  &	96.2  &	88.6  &	81.7  &	57.2  &	79.3  &	85.6 &	96.5  &	84.6  &	\textbf{78.7}  &	86.4  &	80.5  &	70.7  &	\textbf{51.8} &	78.5  &	77.3  &	49.2  &	28.7  &	40.3  &	59.6  &	72.5  &	0.02  \\
             MoIL~\cite{luo2024moil} & 71.7  &	92.8  &	69.6  &	99.1  &	89.8  &	84.2  &	56.0  &	80.5  &	84.7  &	94.9  &	82.3  &	74.6 &	84.1 &	79.5  &	62.7  &	49.1  &	81.6  &	81.5  &	48.4  &	36.9  &	47.5  &	\textbf{60.9}  &	73.0  & 0.63 \\
             RepAdapter$_{attn}$ ~\cite{luo2023towards} & 70.7 & 91.6 &	72.5 &	99.1 &	91.3	 &	88.5	 &	54.2 &	81.1 & 84.1	 &	95.7 &	85.1 &	74.6 &	84.9 &	\textbf{81.6} &	\textbf{69.1} &	50.4 &	81.9 &	79.5 &	45.6 &	34.6 &	41.9 &	60.6 &	73.3  &  0.11 \\
             Adapter+$_{r=1}$ ~\cite{steitz2024adapters} & 85.4  &	92.4 & 	73.1  &	99.1 	 & 91.3  & 	83.1 & 	58.1 & 	83.2 & 	87.2 & 	96.6 & 	85.3 & 	72.6 & 	85.4 & 	80.7 & 	60.6 & 	50.9 & 	79.9 & 	83.3 & 	55.6 & 	27.1 & 	43.0 & 	60.1 & 	74.0 & 0.07 \\
             \rowcolor{gray!20}
            LoRA+AOFT$^{\dag}$   & 74.2  & \textbf{93.4}  & \textbf{77.7}  & \textbf{99.4}  & \textbf{91.5}  & 85.5  &  \textbf{57.3}  &  \textbf{82.0}  &  86.3  &  95.2  &  84.0  &  75.8  &  85.3  &  78.9  &  63.1  &  51.2  &  \textbf{82.6}  &  \textbf{83.6}  &  \textbf{53.9}  &  31.9  &  \textbf{47.3} &  57.3  &  \textbf{74.1}  &  0.08 \\
			\bottomrule
		\end{tabular}
	}
    \vspace{-0.2cm}
	\label{vitb_vtab}
\end{table*}

\subsection{Experimental Comparisons}\label{sec:experiment}
In this section, we integrate our method with various backbones and provide a quantitative comparison to evaluate its effectiveness.

\noindent
\textbf{Comparison with the existing PEFT methods.}
We evaluate the AOFT integrate state-of-the-art PEFT approaches on two benchmarks: FGVC and VTAB-1k. When applied to Adapter, VPT, LoRA, OFT and GOFT. More orthogonal fine-tuning practices have been conducted on language models, so we do not compare with them. Our method consistently demonstrates significant performance improvements, as shown in Tables~\cref{vitb_fgvc} and~\cref{vitb_vtab}. Notably, these enhancements are achieved while reducing the parameter count by more than half. In particular, on the VTAB-1k~\cite{zhai2019visual} dataset, integrating AOFT with the VPT approach results in an average performance gain of 1.3\% across 19 datasets, while simultaneously reducing the parameter count by an order of magnitude. Furthermore, for the LoRA method, AOFT$^{\dag}$ achieves 1.8\% improvement in performance. These results demonstrate the potential of AOFT to facilitate the fine-tuning of foundational models under resource-constrained conditions.
\begin{table*}[!tb]
	\centering
        \caption{
        Performance comparison on VTAB-1k using ViT-Large and ViT-Huge pre-trained on ImageNet-21k as the backbone. The notation "$(\cdot)$" indicates the number of tasks in the subgroup. Detailed results are presented in Appendix~\ref{sec:LHSdetails}. The best results are shown in \textbf{bold}.
        }
        \vspace{-0.2cm}
	  \resizebox{\linewidth}{!}{
		\begin{tabular}{c|ccc|cc|ccc|cc}
			\toprule
			\multicolumn{1}{c|}{\multirow{2}[2]{*}{\diagbox{\textbf{Methods}}{\textbf{Datasets}}}} & \multicolumn{5}{c|}{(a) ViT-Large} & \multicolumn{5}{c}{(b) ViT-Huge} \\
			& \textbf{Natural (7)} & \textbf{Specialized (4)} & \multicolumn{1}{c}{\textbf{Structed (8)}} & \textbf{Mean} & \textbf{Params.(M)} & \textbf{Natural (7)} & \textbf{Specialized (4)} & \multicolumn{1}{c}{\textbf{Structed (8)}} & \textbf{Mean} & \textbf{Params.(M)} \\
			\midrule
			Full fine-tuning & 74.7  & 83.8  & 48.1  & 65.4  & 303.40  & 70.9  & 83.6  & 46.0  & 63.1  & 630.90  \\
			Linear probing & 70.9  & 69.1  & 25.8  & 51.5  & 0.05  & 67.9  & 79.0  & 26.1  & 52.7  & 0.06  \\
			\midrule
			Adapter~\cite{houlsby2019parameter} & 68.6  & 73.5  & 29.0  & 52.9  & 2.38  & 68.1  & 76.4  & 24.5  & 51.5  & 5.78  \\
            Adapter+AOFT*  & \textbf{70.7}  & \textbf{77.0}  & \textbf{44.4}  & \textbf{60.9}  & 0.10  & \textbf{77.7}  & \textbf{81.8}  & \textbf{37.1}  & \textbf{61.5}  & 0.17  \\
            \midrule
			LoRA~\cite{hu2021lora}  & 81.4  & 85.0  & 57.3  & 72.0  & 0.74  & 77.1  & 83.5  & 55.4  & 69.3  & 1.21  \\
            LoRA+AOFT*   & \textbf{83.3}  & \textbf{85.9}  & \textbf{60.2}  & \textbf{74.3}  & 0.15  &  \textbf{78.8} & \textbf{83.8}  & \textbf{58.3}  & \textbf{71.3}  & 0.20  \\
			\bottomrule
		\end{tabular}%
	}
	\vspace{-0.4cm}
    \label{tab:wo_vpt_vitlh}
\end{table*}


\begin{table}[!tb]
	\centering
        \caption{Performance comparison on VTAB-1k using Swin Transformer pre-trained on ImageNet-21k as the backbone. "$(\cdot)$" indicates the number of tasks in the subgroup. Detailed results are presented in Appendix~\ref{sec:LHSdetails}. The best results are shown in \textbf{bold}.}
        \vspace{-0.2cm}
	\resizebox{\linewidth}{!}{
		\begin{tabular}{c|c|c|c|cc}
			\toprule
			\diagbox{\textbf{Methods}}{\textbf{Datasets}} & \textbf{Natural (7)} & \textbf{Specialized (4)} & \textbf{Structed (8)} & \textbf{Mean Total} & \textbf{Params.(M)} \\
			\midrule
			Full fine-tuning & 79.1  & 86.2  & 59.7  & 72.4  & 86.80  \\
			Linear probing & 73.5  & 80.8  & 33.5  & 58.2  & 0.05  \\
			\midrule
			MLP-4~\cite{jia2022visual} & 70.6  & 80.7  & 31.2  & 57.7  & 4.04  \\
			Partial~\cite{jia2022visual} & 73.1  & 81.7  & 35.0  & 58.9  & 12.65  \\
			Bias~\cite{zaken2022bitfit}  & 74.2  & 80.1  & 42.4  & 62.1  & 0.25  \\
			VPT-Shallow~\cite{jia2022visual} & 79.9  & 82.5  & 37.8  & 62.9  & 0.05  \\
			VPT-Deep~\cite{jia2022visual} & 76.8  & 84.5  & 53.4  & 67.7  & 0.22  \\
			ARC~\cite{dong2023efficient}   & 79.0  & 86.6  & 59.9 & 72.6  & 0.27  \\
			RLRR~\cite{dong2024low}   & 81.3 & 86.7 & 59.0  & 73.0 & 0.41 \\
            \midrule
            LoRA+AOFT*   & \textbf{82.3} & \textbf{86.8} & \textbf{60.6}  & \textbf{73.3} & 0.14 \\
			\bottomrule
		\end{tabular}%
        }
	\label{tab:swinb_vtab}%
    \vspace{-0.6cm}
\end{table}%

\noindent
\textbf{Experiments on larger-scale ViT backbones.}
In addition to evaluating our method on the ViT-B backbone, we further conducted experiments on larger backbones, ViT-L and ViT-H~\cite{dosovitskiy2020image}, to assess the effectiveness of AOFT when applied to more complex and large-scale architectures. As shown in~\cref{tab:wo_vpt_vitlh}, models integrated with AOFT consistently outperform their respective baselines, demonstrating that AOFT maintains strong performance even when applied to larger backbone networks.

\noindent
\textbf{Experiments on hierarchical Vision Transformers.}
To further explore the scalability of AOFT, we extend our method to the Swin Transformer~\cite{liu2021swin}, a hierarchical vision model that partitions the input into multiple stages, with each stage consisting of transformer modules operating at a fixed and unique feature resolution. As shown in~\cref{tab:swinb_vtab}, the integration of AOFT with LoRA achieves state-of-the-art performance. Compared to ARC, which shares a low-rank matrix, our method reduces the parameter count by half while achieving 0.7\% performance improvement, demonstrating its robustness in adapting to foundation models.

\subsection{Ablation Studies}\label{sec:ablation}
In this section, we conduct a detailed ablation study to investigate the underlying mechanisms of AOFT, examining its impact across different model positions and the effect of varying bottleneck dimensionality on performance across the selected benchmarks.

\noindent
\textbf{Effect of combining AOFT and Low-Rank adaptation.}
To evaluate the adaptive effects of AOFT across different model components, we applied it to various positions, including multiple components within MHA and FFN. Our findings align with the conclusions from the original LoRA paper, where applying LoRA to only the query $\mathbf{W}_q$ and value $\mathbf{W}_v$ matrices yields performance comparable to applying it to all components. Similarly, AOFT achieves competitive results under the same setting when adapters are applied to the feedforward neural network $\mathbf{W}_{\rm{FC1}}$ and $\mathbf{W}_{\rm{FC2}}$, further validating its effectiveness in low-rank adaptation. This demonstrates that AOFT can be seamlessly incorporated into LoRA, further enhancing its adaptability and efficiency in low-rank adaptation.

\begin{figure}[!tb]
	\centering
    \includegraphics[width=0.85\linewidth]{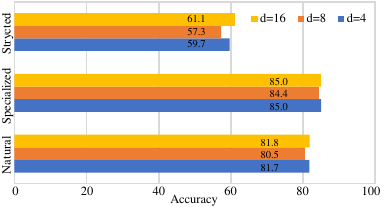}
    \vspace{-0.2cm}
    \caption{Ablation study on the impact of different bottleneck dimensions of adaptive matrices in AOFT. The bar chart represents the Top-1 Test Accuracy. Each bar represents the average accuracy of predictions for that particular type of dataset.}
    \label{fig:bottleneck}
    \vspace{-0.2cm}
\end{figure}

\noindent
\textbf{Effect of bottleneck dimension on the performance of AOFT.}
Fig.~\ref{fig:bottleneck} illustrates the impact of bottleneck dimensionality on the performance of AOFT within low-rank structures. Previous studies~\cite{schmidhuber2015deep, vasuki2017deep, goldberg2017neural, shi2018adaptive} have shown that different tasks require varying matrix dimensions. To systematically investigate this effect, we varied the bottleneck dimension in our experiments. Notably, our approach constructs matrices from a single vector, enabling the generation of an arbitrary number of approximately orthogonal column vectors. This property allows for flexible adjustment of dimensionality without increasing the total number of parameters. As observed in the figure, AOFT demonstrates significant potential, highlighting its advantages and broader applicability. This suggests that AOFT can effectively adapt to different task requirements while maintaining parameter efficiency, making it a promising approach for low-rank adaptation.


\begin{table}[!tb]
	\centering
        \caption{AOFT is used to replace the down/up-projection matrices, and the experiment was conducted on ViT-B/16 backbone and VTAB-1k dataset. According to previous experiments, AOFT is applied to the $\{\mathbf{W}_q,\mathbf{W}_v\}$ and $\{\mathbf{W}_\mathrm{FC1}, \mathbf{W}_\mathrm{FC2}\}$ matrices, and different bottleneck dimensions were set for ablation experiments. The best results are shown in \textbf{bold}.}
        \vspace{-0.2cm}
	\resizebox{\linewidth}{!}{
		\begin{tabular}{c|c|c|c|cc}
			\toprule
			\diagbox{\textbf{Methods}}{\textbf{Datasets}} & \textbf{Natural (7)} & \textbf{Specialized (4)} & \textbf{Structed (8)} & \textbf{Mean Total} & \textbf{Params.(M)} \\
			\midrule
			LoRA$(\mathbf{W}_q,\mathbf{W}_v)$ & 79.5 & 84.6 & 59.8 & 72.3 &  0.29 \\
            LoRA+AOFT$(\mathbf{W}_q,\mathbf{W}_v)$ & \textbf{81.7} & 85.0 & 59.7 & 73.1 &  0.08 \\
            LoRA+AOFT$(\mathbf{W}_q,\mathbf{W}_v,\mathbf{W}_\mathrm{FC1},\mathbf{W}_\mathrm{FC2})$ & 81.4 & \textbf{85.3} & \textbf{60.6} & \textbf{73.4} & 0.16  \\
            \midrule
            Adapter$(\mathbf{W}_\mathrm{FFN})$ & 79.0 & 84.1 & 58.5 & 71.4 & 0.16 \\
            Adapter+AOFT$(\mathbf{W}_\mathrm{FFN})$ & \textbf{81.5} & 85.2 & 59.5 & 73.0 &  0.06 \\
            Adapter+AOFT$(\mathbf{W}_\mathrm{FFN},\mathbf{W}_\mathrm{MHA})$ & 80.7 & \textbf{86.2} & \textbf{60.9} & \textbf{73.5} & 0.08 \\
        \bottomrule
		\end{tabular}%
	}
    \vspace{-0.2cm}
	\label{tab:att_ffn}%
\end{table}%

\begin{table}[!tb]
	\centering
        \caption{Comparison of the L2-norms of prompt matrices between VPT-Deep and VPT-Deep+AOFT. The best results are shown in \textbf{bold}.}
        \vspace{-0.2cm}
	\resizebox{\linewidth}{!}{
		\begin{tabular}{c|c|c|c|c|c|c}
			\toprule
			\diagbox{\textbf{Methods}}{\textbf{Layer}} & \textbf{Layer 1} & \textbf{Layer 2} & \textbf{Layer 3} & \textbf{Layer 4} & \textbf{Layer 5} & \textbf{Layer 6} \\
			\midrule
			VPT-Deep~\cite{jia2022visual} & 7.41 & 6.17 & 6.02 & 5.08 & 5.36 & 5.01 \\
            VPT-Deep+AOFT & 1.09 & 1.04 & 1.07 & 1.04 & 1.04 & 1.05 \\
            \toprule
			\diagbox{\textbf{Methods}}{\textbf{Layer}} & \textbf{Layer 7} & \textbf{Layer 8} & \textbf{Layer 9} & \textbf{Layer 10} & \textbf{Layer 11} & \textbf{Layer 12} \\
			\midrule
			VPT-Deep~\cite{jia2022visual} & 4.28 & 3.77 & 3.09 & 2.67 & 2.15 & 3.03 \\
            VPT-Deep+AOFT & 1.04 & 1.03 & 1.10 & 1.04 & 1.04 & 1.16 \\
        \bottomrule
		\end{tabular}%
	}
	\label{tab:vpt-L2}%
    \vspace{-0.6cm}
\end{table}%

\begin{figure}[!tb]
    \centering
    \begin{subfigure}[t]{0.22\textwidth}
        \includegraphics[width=\textwidth]{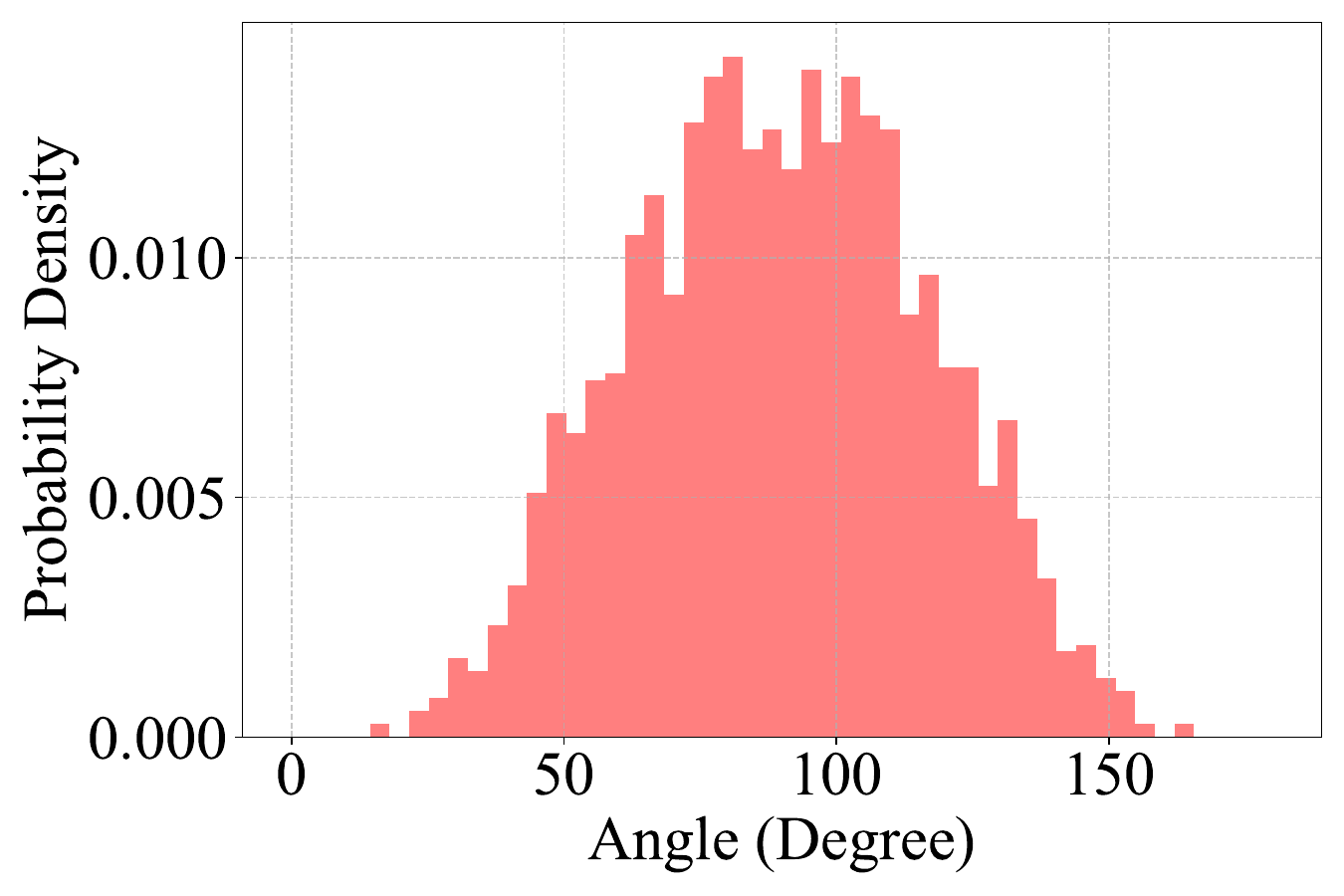} 
        \caption{The angle distribution of any two column vectors of prompt matrix in VPT-Shallow.}
        \label{fig:vpt-shallow-base}
    \end{subfigure}
    \begin{subfigure}[t]{0.22\textwidth}
        \includegraphics[width=\textwidth]{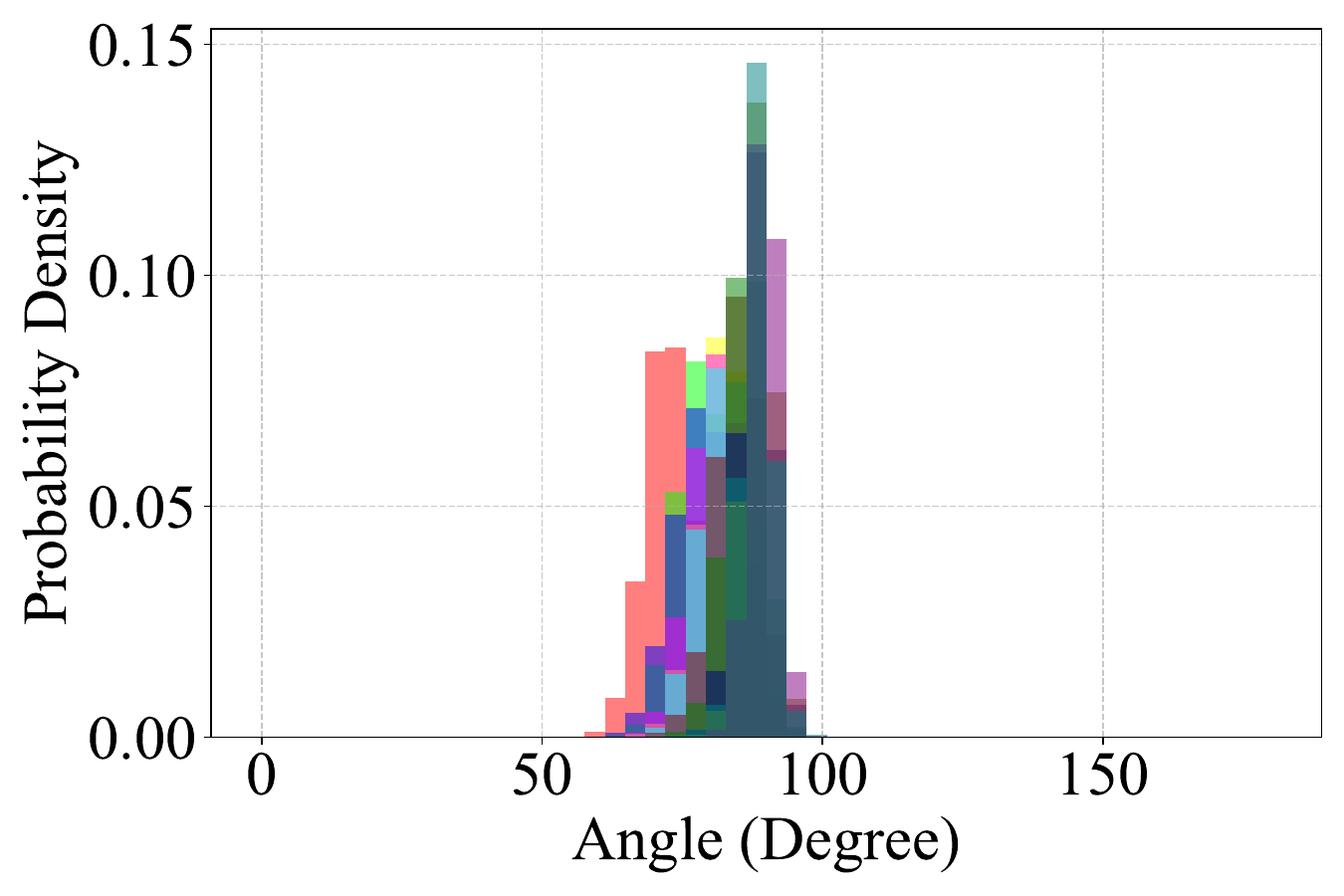} 
        \caption{The angle distribution of any two column vectors of prompt matrices in VPT-Deep.}
        \label{fig:vpt-deep-base}
    \end{subfigure}
    \begin{subfigure}[t]{0.22\textwidth}
        \includegraphics[width=\textwidth]{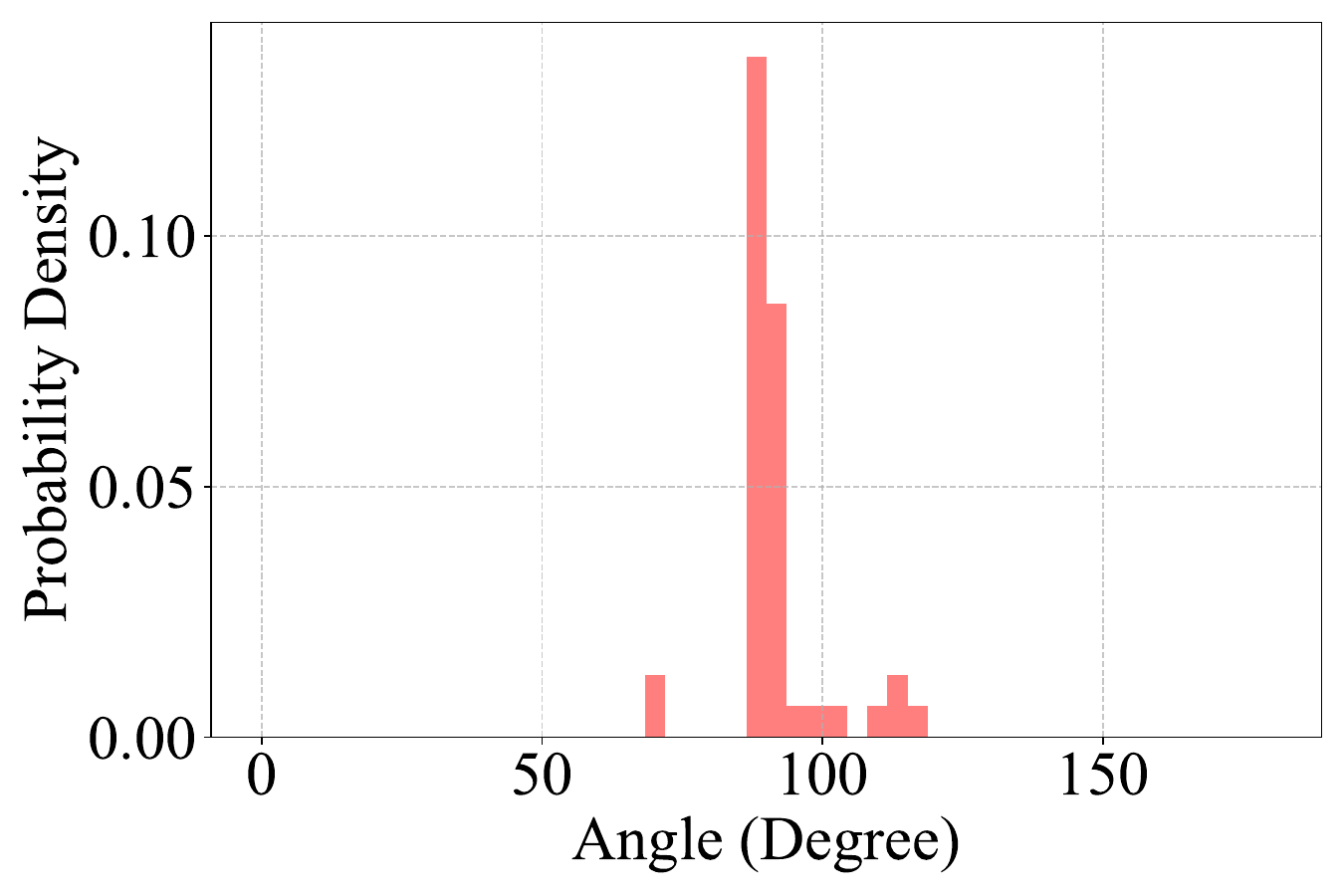} 
        \caption{The angle distribution of any two column vectors of matrix in VPT-shallow with AOFT.}
        \label{fig:vpt-shallow-ours}
    \end{subfigure}
    \begin{subfigure}[t]{0.22\textwidth}
        \includegraphics[width=\textwidth]{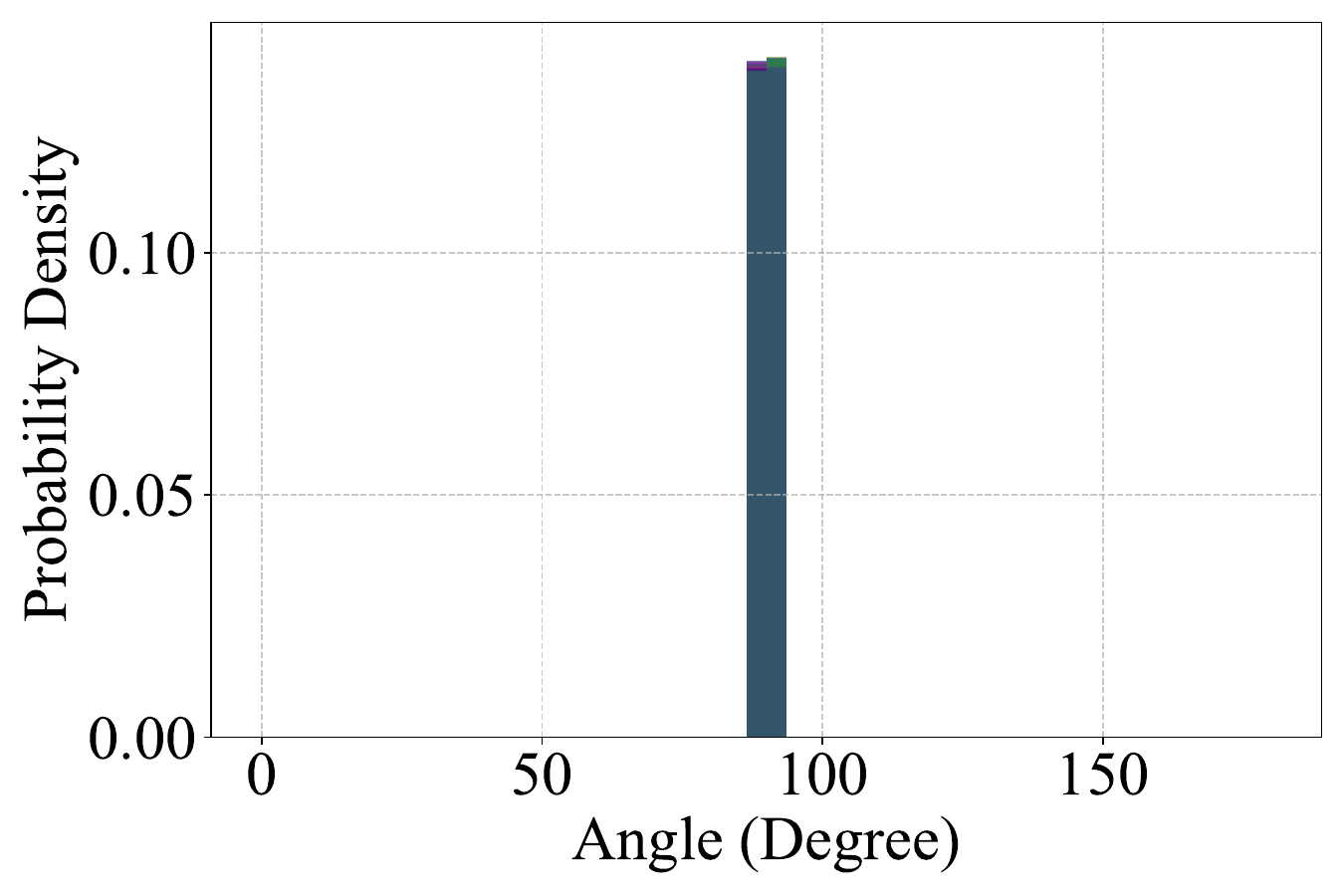} 
        \caption{The angle distribution of any two column vectors of prompt matrices in VPT-deep with AOFT.}
        \label{fig:vpt-deep-ours}
    \end{subfigure}
    \vspace{-0.2cm}
    \caption{Comparison of the angle distribution between any two column vectors within the prompt matrices of VPT and those enhanced with AOFT. These results are all obtained from models trained on the dtd dataset. Their legends can be found in Appendix~\ref{sec:Legend}.}
    \label{fig:vpt-ours}
    \vspace{-0.5cm}
\end{figure}

\noindent
\textbf{The L2-norms and angle distribution of VPT integrated with AOFT.}
In Eq.(\ref{eq:RC_L2}), we provide a general derivation of the impact of the L2-norm on generalization, which suggests that larger norms lead to poorer generalization performance. Furthermore, the experiments in Section~\ref{sec:experiment} have already demonstrated that our method effectively improves performance when applied to VPT. The result is show in Tab.~\ref{vitb_vtab}. To further investigate this effect, we analyze the behavior of the L2-norm across different layers to illustrate the influence of AOFT on existing PEFT methods, with the results presented in~\cref{tab:vpt-L2}. Additionally, Fig.~\ref{fig:vpt-ours} shown the angular distribution between arbitrary column vectors within the prompt matrix when applying AOFT to VPT. These findings indicate that the approximate orthogonality strategy helps mitigate generalization errors, thereby improving overall model performance.

\section{Conclusions}
In this work, we propose the Approximately Orthogonal Fine-Tuning (AOFT) strategy, which generates approximately orthogonal vectors for down/up-projection matrices using a single learnable vector, aligning their properties with the backbone. Specifically, we replace the down- and up-projection matrices in prevalent methods like LoRA and Adapter with matrices that have approximately orthogonal columns. This strategy is linked to reduced generalization error and enhanced model capability. Experimental results across various image classification downstream tasks confirm the efficacy of AOFT in enhancing generalization capability, achieving competitive performance.

\section{Acknowledgments and Disclosure of Funding}
This work was supported by Major Science and Technology Innovation Special Project Plan of Xianyang (Program No. L2024-ZDKJ-ZDGG-GY-0010) and Natural Science Basic Research Program of Shaanxi (Program No.2024JC-YBMS-464).

{
    \normalem
    \small
    \bibliographystyle{ieeenat_fullname}
    \bibliography{main}

\begin{thebibliography}{34}
\providecommand{\natexlab}[1]{#1}
\providecommand{\url}[1]{\texttt{#1}}
\expandafter\ifx\csname urlstyle\endcsname\relax
  \providecommand{\doi}[1]{doi: #1}\else
  \providecommand{\doi}{doi: \begingroup \urlstyle{rm}\Url}\fi

\bibitem[Dettmers et~al.(2023)Dettmers, Pagnoni, Holtzman, and
  Zettlemoyer]{dettmers2023qlora}
Tim Dettmers, Artidoro Pagnoni, Ari Holtzman, and Luke Zettlemoyer.
\newblock Qlora: Efficient finetuning of quantized llms.
\newblock \emph{Advances in neural information processing systems},
  36:\penalty0 10088--10115, 2023.

\bibitem[Dong et~al.(2023)Dong, Yan, Lin, and Wang]{dong2023efficient}
Wei Dong, Dawei Yan, Zhijun Lin, and Peng Wang.
\newblock Efficient adaptation of large vision transformer via adapter
  re-composing.
\newblock In \emph{Thirty-seventh Conference on Neural Information Processing
  Systems}, 2023.

\bibitem[Dong et~al.(2024)Dong, Zhang, Chen, Yan, Lin, Yan, Wang, and
  Yang]{dong2024low}
Wei Dong, Xing Zhang, Bihui Chen, Dawei Yan, Zhijun Lin, Qingsen Yan, Peng
  Wang, and Yang Yang.
\newblock Low-rank rescaled vision transformer fine-tuning: A residual design
  approach.
\newblock In \emph{Proceedings of the IEEE/CVF Conference on Computer Vision
  and Pattern Recognition}, pages 16101--16110, 2024.

\bibitem[Dosovitskiy et~al.(2020)Dosovitskiy, Beyer, Kolesnikov, Weissenborn,
  Zhai, Unterthiner, Dehghani, Minderer, Heigold, Gelly,
  et~al.]{dosovitskiy2020image}
Alexey Dosovitskiy, Lucas Beyer, Alexander Kolesnikov, Dirk Weissenborn,
  Xiaohua Zhai, Thomas Unterthiner, Mostafa Dehghani, Matthias Minderer, Georg
  Heigold, Sylvain Gelly, et~al.
\newblock An image is worth 16x16 words: Transformers for image recognition at
  scale.
\newblock In \emph{International Conference on Learning Representations}, 2020.

\bibitem[Gebru et~al.(2017)Gebru, Krause, Wang, Chen, Deng, and
  Fei-Fei]{gebru2017fine}
Timnit Gebru, Jonathan Krause, Yilun Wang, Duyun Chen, Jia Deng, and Li
  Fei-Fei.
\newblock Fine-grained car detection for visual census estimation.
\newblock In \emph{Proceedings of the AAAI Conference on Artificial
  Intelligence}, 2017.

\bibitem[Goar and Yadav(2024)]{goar2024foundations}
Vishal Goar and Nagendra~Singh Yadav.
\newblock Foundations of machine learning.
\newblock In \emph{Intelligent Optimization Techniques for Business Analytics},
  pages 25--48. IGI Global, 2024.

\bibitem[Goldberg(2017)]{goldberg2017neural}
Yoav Goldberg.
\newblock \emph{Neural network methods in natural language processing}.
\newblock Morgan \& Claypool Publishers, 2017.

\bibitem[Golub and Van~Loan(2013)]{golub2013matrix}
Gene~H Golub and Charles~F Van~Loan.
\newblock \emph{Matrix computations}.
\newblock JHU press, 2013.

\bibitem[Houlsby et~al.(2019)Houlsby, Giurgiu, Jastrzebski, Morrone,
  De~Laroussilhe, Gesmundo, Attariyan, and Gelly]{houlsby2019parameter}
Neil Houlsby, Andrei Giurgiu, Stanislaw Jastrzebski, Bruna Morrone, Quentin
  De~Laroussilhe, Andrea Gesmundo, Mona Attariyan, and Sylvain Gelly.
\newblock Parameter-efficient transfer learning for nlp.
\newblock In \emph{International Conference on Machine Learning}, pages
  2790--2799. PMLR, 2019.

\bibitem[Hu et~al.(2021)Hu, Wallis, Allen-Zhu, Li, Wang, Wang, Chen,
  et~al.]{hu2021lora}
Edward~J Hu, Phillip Wallis, Zeyuan Allen-Zhu, Yuanzhi Li, Shean Wang, Lu Wang,
  Weizhu Chen, et~al.
\newblock Lora: Low-rank adaptation of large language models.
\newblock In \emph{International Conference on Learning Representations}, 2021.

\bibitem[Jia et~al.(2022)Jia, Tang, Chen, Cardie, Belongie, Hariharan, and
  Lim]{jia2022visual}
Menglin Jia, Luming Tang, Bor-Chun Chen, Claire Cardie, Serge Belongie, Bharath
  Hariharan, and Ser-Nam Lim.
\newblock Visual prompt tuning.
\newblock In \emph{European Conference on Computer Vision}, pages 709--727.
  Springer, 2022.

\bibitem[Jie and Deng(2023)]{jie2023fact}
Shibo Jie and Zhi-Hong Deng.
\newblock Fact: Factor-tuning for lightweight adaptation on vision transformer.
\newblock In \emph{Proceedings of the AAAI Conference on Artificial
  Intelligence}, pages 1060--1068, 2023.

\bibitem[Khosla et~al.(2011)Khosla, Jayadevaprakash, Yao, and
  Li]{khosla2011novel}
Aditya Khosla, Nityananda Jayadevaprakash, Bangpeng Yao, and Fei-Fei Li.
\newblock Novel dataset for fine-grained image categorization: Stanford dogs.
\newblock In \emph{Proc. CVPR workshop on fine-grained visual categorization
  (FGVC)}. Citeseer, 2011.

\bibitem[Kopiczko et~al.(2023)Kopiczko, Blankevoort, and
  Asano]{kopiczko2023vera}
Dawid~J Kopiczko, Tijmen Blankevoort, and Yuki~M Asano.
\newblock Vera: Vector-based random matrix adaptation.
\newblock \emph{arXiv preprint arXiv:2310.11454}, 2023.

\bibitem[Lian et~al.(2022)Lian, Zhou, Feng, and Wang]{lian2022scaling}
Dongze Lian, Daquan Zhou, Jiashi Feng, and Xinchao Wang.
\newblock Scaling \& shifting your features: A new baseline for efficient model
  tuning.
\newblock \emph{Advances in Neural Information Processing Systems},
  35:\penalty0 109--123, 2022.

\bibitem[Liu et~al.(2024)Liu, Wang, Yin, Molchanov, Wang, Cheng, and
  Chen]{liu2024dora}
Shih-Yang Liu, Chien-Yi Wang, Hongxu Yin, Pavlo Molchanov, Yu-Chiang~Frank
  Wang, Kwang-Ting Cheng, and Min-Hung Chen.
\newblock Dora: Weight-decomposed low-rank adaptation.
\newblock In \emph{Forty-first International Conference on Machine Learning},
  2024.

\bibitem[Liu et~al.(2023)Liu, Qiu, Feng, Xiu, Xue, Yu, Feng, Liu, Heo, Peng,
  et~al.]{liu2023parameter}
Weiyang Liu, Zeju Qiu, Yao Feng, Yuliang Xiu, Yuxuan Xue, Longhui Yu, Haiwen
  Feng, Zhen Liu, Juyeon Heo, Songyou Peng, et~al.
\newblock Parameter-efficient orthogonal finetuning via butterfly
  factorization.
\newblock \emph{arXiv preprint arXiv:2311.06243}, 2023.

\bibitem[Liu et~al.(2021)Liu, Lin, Cao, Hu, Wei, Zhang, Lin, and
  Guo]{liu2021swin}
Ze Liu, Yutong Lin, Yue Cao, Han Hu, Yixuan Wei, Zheng Zhang, Stephen Lin, and
  Baining Guo.
\newblock Swin transformer: Hierarchical vision transformer using shifted
  windows.
\newblock In \emph{Proceedings of the IEEE/CVF international conference on
  computer vision}, pages 10012--10022, 2021.

\bibitem[Luo et~al.(2023)Luo, Huang, Zhou, Sun, Jiang, Wang, and
  Ji]{luo2023towards}
Gen Luo, Minglang Huang, Yiyi Zhou, Xiaoshuai Sun, Guannan Jiang, Zhiyu Wang,
  and Rongrong Ji.
\newblock Towards efficient visual adaption via structural re-parameterization.
\newblock \emph{arXiv preprint arXiv:2302.08106}, 2023.

\bibitem[Luo et~al.(2024)Luo, Zhou, Huang, Ren, Sun, and Ji]{luo2024moil}
Gen Luo, Yiyi Zhou, Minglang Huang, Tianhe Ren, Xiaoshuai Sun, and Rongrong Ji.
\newblock Moil: Momentum imitation learning for efficient vision-language
  adaptation.
\newblock \emph{IEEE Transactions on Pattern Analysis and Machine
  Intelligence}, 2024.

\bibitem[Ma et~al.(2024)Ma, Chu, Yang, Lin, Gao, and Zhao]{ma2024parameter}
Xinyu Ma, Xu Chu, Zhibang Yang, Yang Lin, Xin Gao, and Junfeng Zhao.
\newblock Parameter efficient quasi-orthogonal fine-tuning via givens rotation.
\newblock \emph{arXiv preprint arXiv:2404.04316}, 2024.

\bibitem[Mayer(2013)]{mayer2013simple}
Istv{\'a}n Mayer.
\newblock \emph{Simple theorems, proofs, and derivations in quantum chemistry}.
\newblock Springer Science \& Business Media, 2013.

\bibitem[Nilsback and Zisserman(2008)]{nilsback2008automated}
Maria-Elena Nilsback and Andrew Zisserman.
\newblock Automated flower classification over a large number of classes.
\newblock In \emph{2008 Sixth Indian conference on computer vision, graphics \&
  image processing}, pages 722--729. IEEE, 2008.

\bibitem[Qiu et~al.(2023)Qiu, Liu, Feng, Xue, Feng, Liu, Zhang, Weller, and
  Sch{\"o}lkopf]{qiu2023controlling}
Zeju Qiu, Weiyang Liu, Haiwen Feng, Yuxuan Xue, Yao Feng, Zhen Liu, Dan Zhang,
  Adrian Weller, and Bernhard Sch{\"o}lkopf.
\newblock Controlling text-to-image diffusion by orthogonal finetuning.
\newblock \emph{Advances in Neural Information Processing Systems},
  36:\penalty0 79320--79362, 2023.

\bibitem[Schmidhuber(2015)]{schmidhuber2015deep}
J{\"u}rgen Schmidhuber.
\newblock Deep learning in neural networks: An overview.
\newblock \emph{Neural networks}, 61:\penalty0 85--117, 2015.

\bibitem[Shi et~al.(2018)Shi, Lin, Hwang, Yang, and Chen]{shi2018adaptive}
Haobin Shi, Zhiqiang Lin, Kao-Shing Hwang, Shike Yang, and Jialin Chen.
\newblock An adaptive strategy selection method with reinforcement learning for
  robotic soccer games.
\newblock \emph{IEEE Access}, 6:\penalty0 8376--8386, 2018.

\bibitem[Steitz and Roth(2024)]{steitz2024adapters}
Jan-Martin~O Steitz and Stefan Roth.
\newblock Adapters strike back.
\newblock In \emph{Proceedings of the IEEE/CVF Conference on Computer Vision
  and Pattern Recognition}, pages 23449--23459, 2024.

\bibitem[Trefethen and Bau(2022)]{trefethen2022numerical}
Lloyd~N Trefethen and David Bau.
\newblock \emph{Numerical linear algebra}.
\newblock SIAM, 2022.

\bibitem[Van~Horn et~al.(2015)Van~Horn, Branson, Farrell, Haber, Barry,
  Ipeirotis, Perona, and Belongie]{van2015building}
Grant Van~Horn, Steve Branson, Ryan Farrell, Scott Haber, Jessie Barry, Panos
  Ipeirotis, Pietro Perona, and Serge Belongie.
\newblock Building a bird recognition app and large scale dataset with citizen
  scientists: The fine print in fine-grained dataset collection.
\newblock In \emph{Proceedings of the IEEE conference on computer vision and
  pattern recognition}, pages 595--604, 2015.

\bibitem[Vasuki and Govindaraju(2017)]{vasuki2017deep}
A Vasuki and S Govindaraju.
\newblock Deep neural networks for image classification.
\newblock In \emph{Deep Learning for Image Processing Applications}, pages
  27--49. IOS Press, 2017.

\bibitem[Wah et~al.(2011)Wah, Branson, Welinder, Perona, and
  Belongie]{wah2011caltech}
Catherine Wah, Steve Branson, Peter Welinder, Pietro Perona, and Serge
  Belongie.
\newblock The caltech-ucsd birds-200-2011 dataset.
\newblock 2011.

\bibitem[Zaken et~al.(2022)Zaken, Goldberg, and Ravfogel]{zaken2022bitfit}
Elad~Ben Zaken, Yoav Goldberg, and Shauli Ravfogel.
\newblock Bitfit: Simple parameter-efficient fine-tuning for transformer-based
  masked language-models.
\newblock In \emph{Proceedings of the 60th Annual Meeting of the Association
  for Computational Linguistics (Volume 2: Short Papers)}, pages 1--9, 2022.

\bibitem[Zhai et~al.(2019{\natexlab{a}})Zhai, Puigcerver, Kolesnikov, Ruyssen,
  Riquelme, Lucic, Djolonga, Pinto, Neumann, Dosovitskiy,
  et~al.]{zhai2019large}
Xiaohua Zhai, Joan Puigcerver, Alexander Kolesnikov, Pierre Ruyssen, Carlos
  Riquelme, Mario Lucic, Josip Djolonga, Andre~Susano Pinto, Maxim Neumann,
  Alexey Dosovitskiy, et~al.
\newblock A large-scale study of representation learning with the visual task
  adaptation benchmark.
\newblock \emph{arXiv preprint arXiv:1910.04867}, 2019{\natexlab{a}}.

\bibitem[Zhai et~al.(2019{\natexlab{b}})Zhai, Puigcerver, Kolesnikov, Ruyssen,
  Riquelme, Lucic, Djolonga, Pinto, Neumann, Dosovitskiy,
  et~al.]{zhai2019visual}
Xiaohua Zhai, Joan Puigcerver, Alexander Kolesnikov, Pierre Ruyssen, Carlos
  Riquelme, Mario Lucic, Josip Djolonga, Andre~Susano Pinto, Maxim Neumann,
  Alexey Dosovitskiy, et~al.
\newblock The visual task adaptation benchmark.
\newblock 2019{\natexlab{b}}.

\end{thebibliography}
}
\clearpage
\setcounter{page}{1}
\setcounter{section}{0}
\maketitlesupplementary

\setcounter{table}{0}
\begin{appendix}

\section{Angle Distribution}\label{sec:Legend}

\begin{figure}[!tbhp]
    \centering
    \begin{subfigure}[t]{0.14\textwidth}
        \includegraphics[width=\textwidth]{figure/Vit_pre/query_angle_distribution.pdf}
        \caption{The degree distribution of pre-trained matrices $\mathbf{W}_{q}$.}
        \label{fig:Avit-query-pre-train}
    \end{subfigure}
    \begin{subfigure}[t]{0.14\textwidth}
        \includegraphics[width=\textwidth]{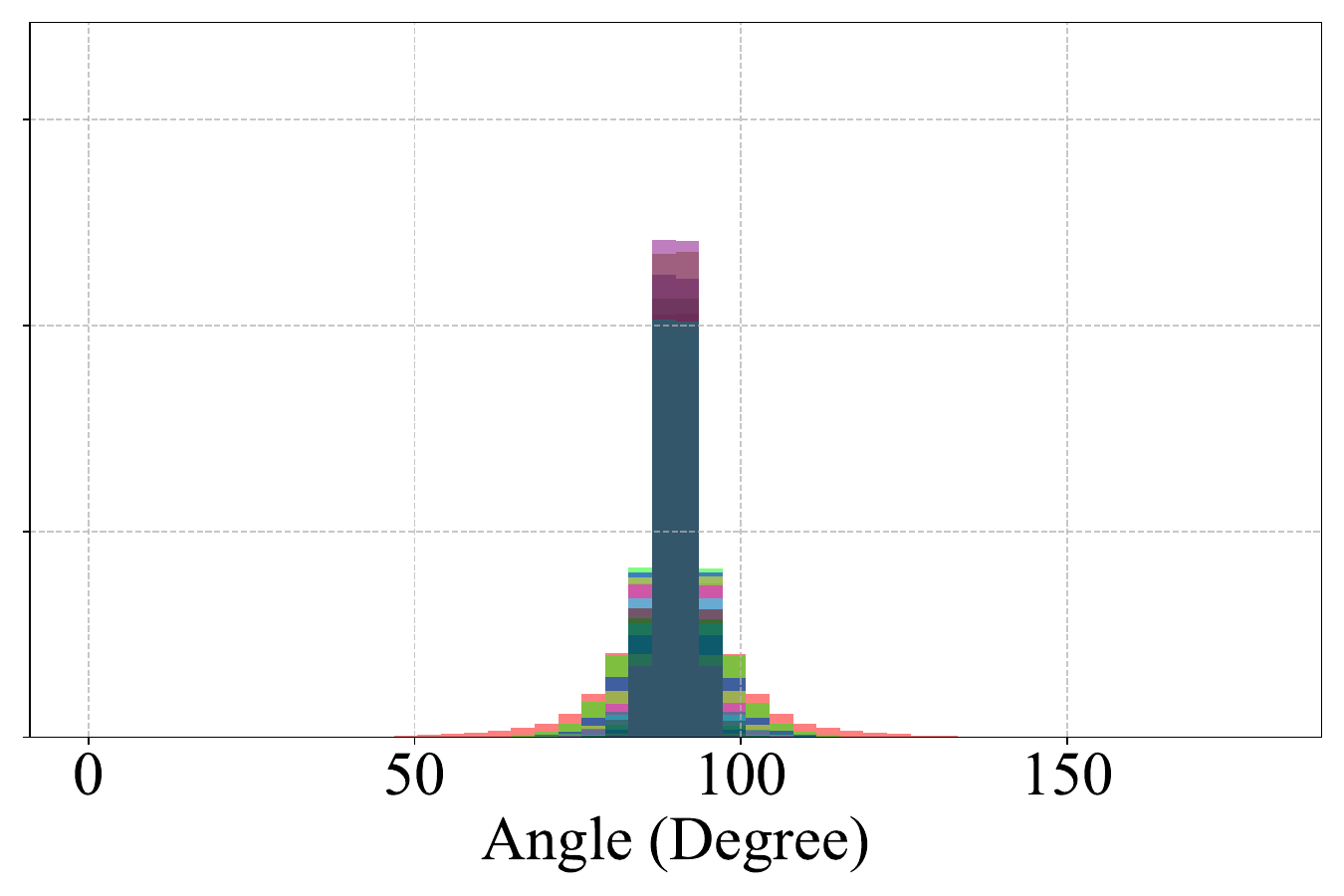}
        \caption{The degree distribution of pre-trained matrices $\mathbf{W}_{k}$.}
        \label{fig:Avit-key-pre-train}
    \end{subfigure}
    \begin{subfigure}[t]{0.14\textwidth}
        \includegraphics[width=\textwidth]{figure/Vit_pre/value_angle_distribution.pdf}
        \caption{The degree distribution of pre-trained matrices $\mathbf{W}_{v}$.}
        \label{fig:Avit-value-pre-train}
    \end{subfigure}
    \begin{subfigure}[t]{0.14\textwidth}
        \includegraphics[width=\textwidth]{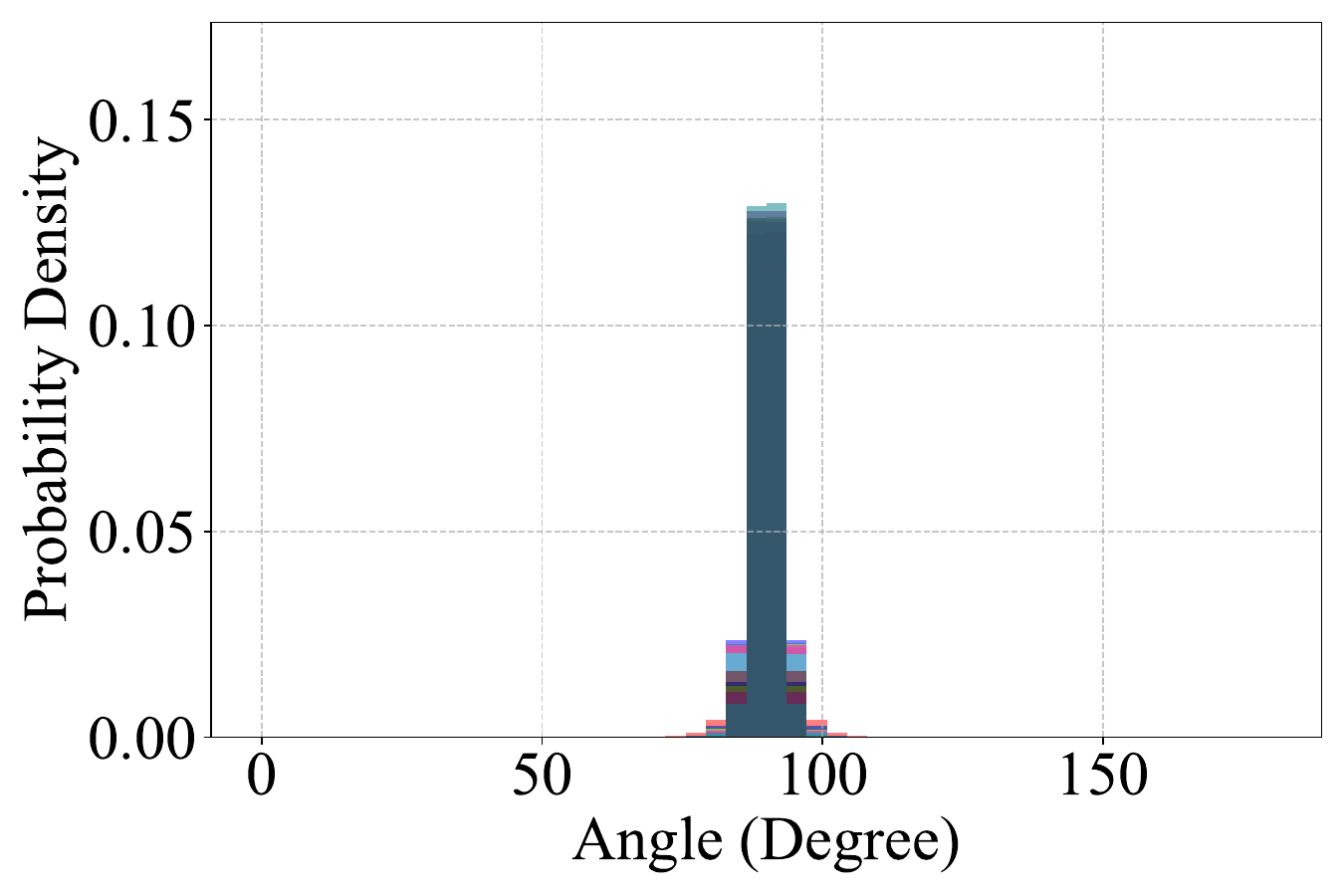}
        \caption{The degree distribution of pre-trained matrices $\mathbf{W}_{o}$.}
        \label{fig:Avit-out-pre-train}
    \end{subfigure}
    \begin{subfigure}[t]{0.14\textwidth}
        \includegraphics[width=\textwidth]{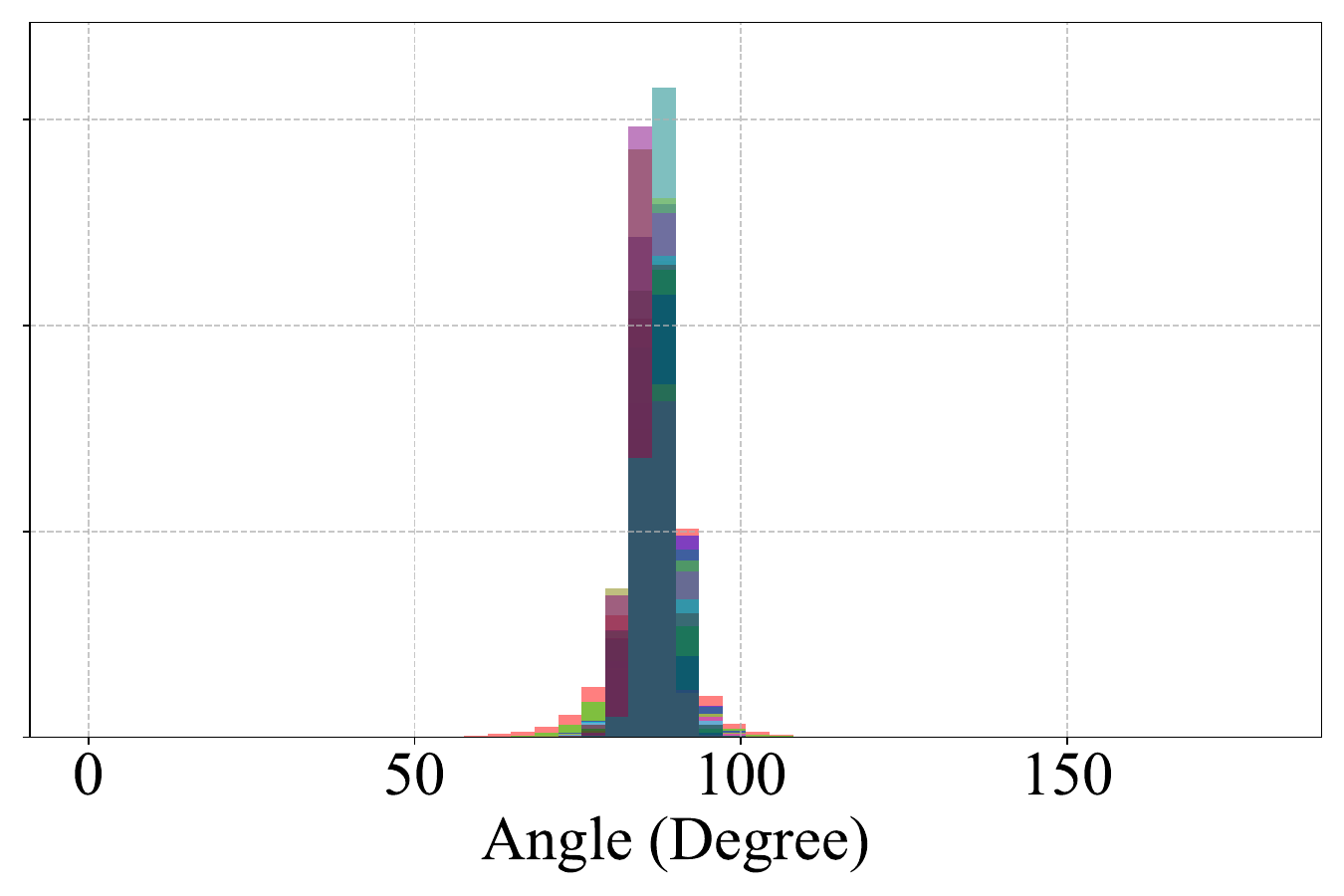}
        \caption{The degree distribution of pre-trained matrices $\mathbf{W}_{\mathrm {FC1}}$.}
        \label{fig:Avit-fc1-pre-train}
    \end{subfigure}
    \begin{subfigure}[t]{0.14\textwidth}
        \includegraphics[width=\textwidth]{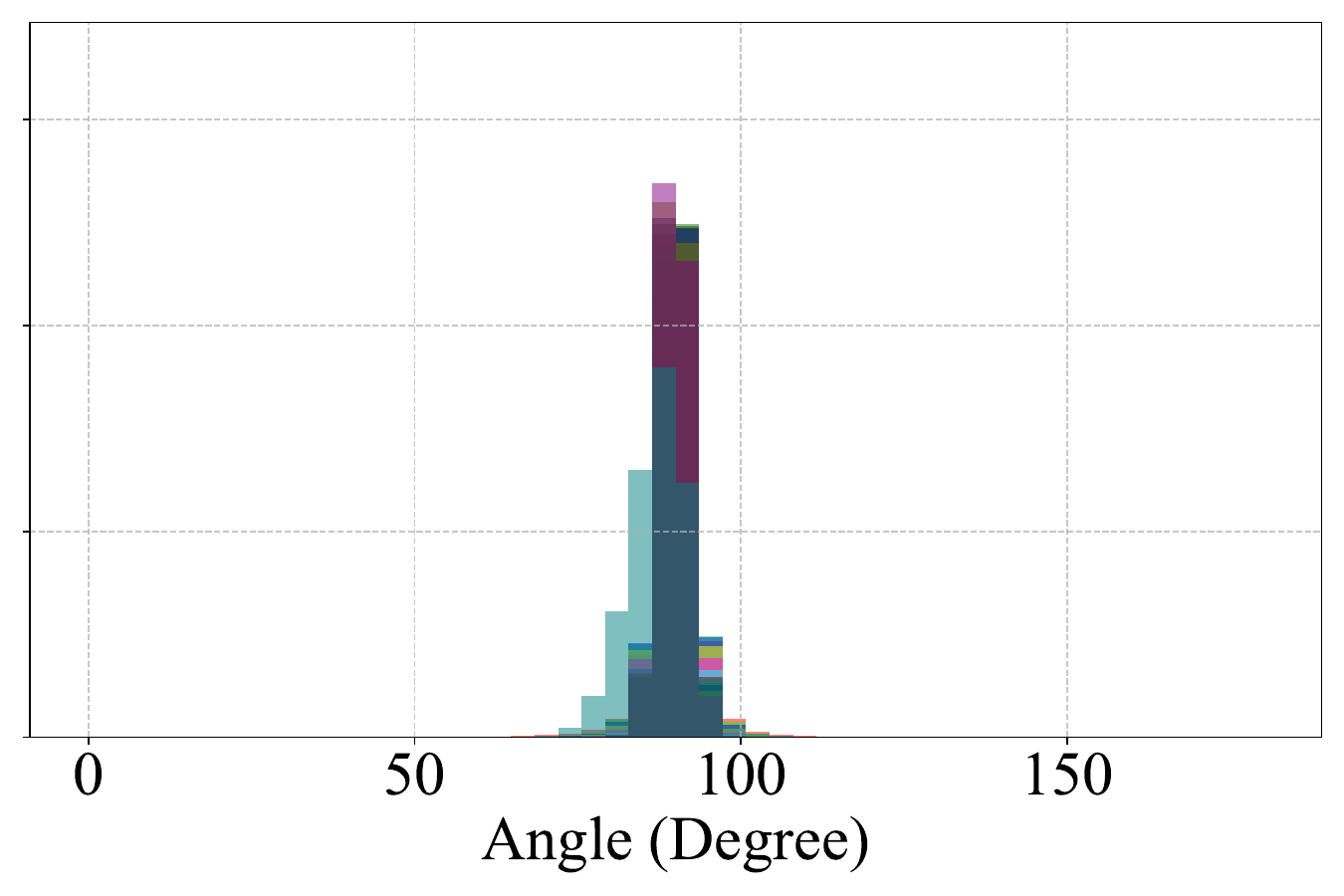}
        \caption{The degree distribution of pre-trained matrices $\mathbf{W}_{\mathrm {FC2}}$.}
        \label{fig:Avit-fc2-pre-train}
    \end{subfigure}
    \caption{Illustration of approximate orthogonality among any two column vectors of weight matrices $\mathbf{W}_{q}, \mathbf{W}_{k}, \mathbf{W}_{v}, \mathbf{W}_{o}, \mathbf{W}_\mathrm{FC1}, \mathbf{W}_\mathrm{FC2}$ in the ViT-B model after training. The histogram represents the distribution of angles between any two column vectors within each weight matrix. Specifically, 
    \subref{fig:Avit-query-pre-train}-\subref{fig:Avit-fc2-pre-train} represent approximate orthogonality in the pre-trained model.Their legends can be found in Fig.~\ref{fig:legend}.}
    \label{fig:Avit-init-pre-train}
\end{figure}

In the weights of the ViT-B/16 pre-trained model $\mathbf{W}_{q}, \mathbf{W}_{k}, \mathbf{W}_{v}, \mathbf{W}_{o}, \mathbf{W}_\mathrm{FC1}, \mathbf{W}_\mathrm{FC2}$. The distribution of pairwise angles between the column vectors in these weight matrices is shown in the ~\cref{fig:Avit-init-pre-train}. It can be observed that in the ViT model, the matrices are approximately orthogonal.

\begin{figure}[!tbhp]
    \centering
    \begin{subfigure}[t]{0.13\textwidth}
        \includegraphics[width=\textwidth]{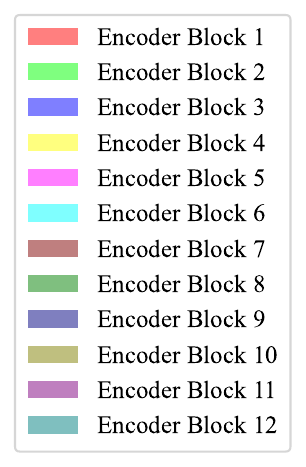} 
        \caption{Legends for Adapter.}
        \label{fig:12_legend}
    \end{subfigure}
    \begin{subfigure}[t]{0.32\textwidth}
        \includegraphics[width=\textwidth]{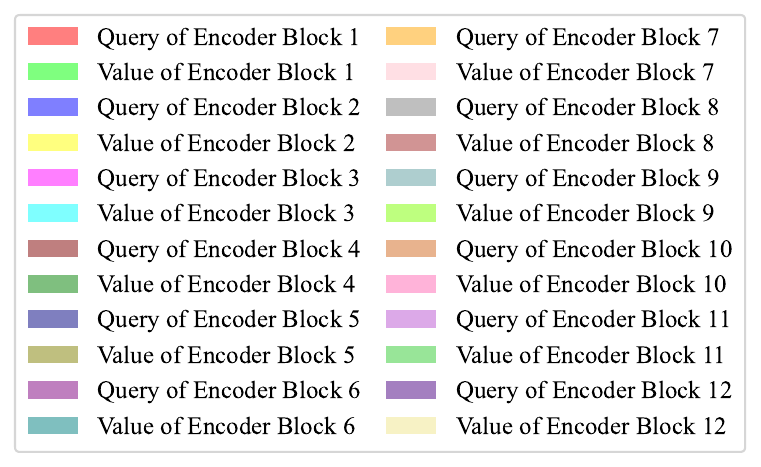} 
        \caption{Legends for LoRA.}
        \label{fig:24_legend}
    \end{subfigure}
    \caption{The legends for the bar chart.}
    \label{fig:legend}
\end{figure}

Fig.~\ref{fig:legend} presents the legends for the statistical charts used throughout this paper. Fig.~\ref{fig:legend}\subref{fig:12_legend}, specifically designed for charts that do not include the LoRA (Low-Rank Adaptation) method, clearly labels the line colors corresponding to different layers of the model, ensuring accurate differentiation among them. Meanwhile, Fig.~\ref{fig:legend}\subref{fig:24_legend} is tailored for charts that incorporate the LoRA method. It distinctly showcases the line colors for different layers and matrices within those layers, facilitating precise differentiation.

\section{Detailed Dataset Statistics}\label{sec:rationale}

In this section, we provide a comprehensive overview of the visual adaptation classification tasks utilized in our study, presenting the specifics for both the Fine-Grained Visual Classification (FGVC) datasets and the Visual Task Adaptation Benchmark-1k (VTAB-1k) datasets. The dataset splits employed in our experiments follow the protocol established by VPT~\cite{jia2022visual}.

\subsection{FGVC Datasets}

The details of the FGVC datasets used in our study are summarized in ~\cref{tab:dataset_fgvc}. For each dataset, we report the number of classes, as well as the sizes of the training, validation, and test sets. These datasets are characterized by their fine-grained nature, requiring models to distinguish between subtle differences in appearance, making them ideal for evaluating the efficacy of visual adaptation techniques.

\subsection{VTAB-1k Datasets}

Similarly, the VTAB-1k datasets used in our experiments are detailed in~\cref{tab:dataset_vtab}. Again, we provide the number of classes and the sizes of the training, validation, and test sets for each dataset. The VTAB-1k benchmark is designed to assess the generalization capabilities of visual models across a diverse range of tasks, making it a valuable tool for evaluating the robustness and adaptability of our proposed methods.

By adhering to the dataset splits established by VPT~\cite{jia2022visual}, we ensure a fair and consistent comparison with existing work, facilitating the reproducibility and validation of our findings.

\begin{table*}[!tbhp] 
  \centering
  \caption{Dataset statistics for FGVC. ``*'' denotes the train/val split of datasets following the dataset setting in VPT~\cite{jia2022visual}.}
    \begin{tabular}{l|l|l|c|c|r}
    \toprule
    \textbf{Dataset} & \textbf{Description} & \textbf{Classes} & \multicolumn{1}{l|}{\textbf{Train size}} & \multicolumn{1}{l|}{\textbf{Val size}} & \textbf{Test size} \\
    \midrule
    CUB-200-2011~\cite{wah2011caltech} & Fine-grained bird species recognition & 200   & \multicolumn{1}{r|}{5,394*} & \multicolumn{1}{r|}{600*} & 5,794 \\
    NABirds~\cite{van2015building} & Fine-grained bird species recognition & 555    & \multicolumn{1}{r|}{21,536*} & \multicolumn{1}{r|}{2,393*} & 24,633 \\
    Oxford Flowers~\cite{nilsback2008automated} & Fine-grained flower species recognition & 102   & \multicolumn{1}{r|}{1,020} & \multicolumn{1}{r|}{1,020} & 6,149 \\
    Stanford Dogs~\cite{khosla2011novel} & Fine-grained dog species recognition & 120   & \multicolumn{1}{r|}{10,800*} & \multicolumn{1}{r|}{1,200*} & 8,580 \\
    Stanford Cars~\cite{gebru2017fine} & Fine-grained car classificatio & 196   & \multicolumn{1}{r|}{7,329*} & \multicolumn{1}{r|}{815*} & 8,041 \\
    \bottomrule
    \end{tabular}%

  \label{tab:dataset_fgvc}
\end{table*}%

\begin{table*}[!tbhp]
  \centering
  \caption{Dataset statistics for VTAB-1k~\cite{zhai2019large}.}
    \begin{tabular}{l|l|r|r|r|r}
    \toprule
    \textbf{Dataset} & \textbf{Description} & \textbf{Classes} & \multicolumn{1}{r|}{\textbf{Train size}} & \multicolumn{1}{r|}{\textbf{Val size}} & \textbf{Test size} \\
    \midrule
    CIFAR-100 
    & \multirow{7}[2]{*}{Natural} & 100   & \multirow{7}[2]{*}{800/1,000} & \multirow{7}[2]{*}{200} & 10,000 \\
    Caltech101 
    &       & 102   &       &       & 6,084 \\
    DTD 
    &       & 47    &       &       & 1,880 \\
    Flowers102 
    &       & 102   &       &       & 6,149 \\
    Pets 
    &       & 37    &       &       & 3,669 \\
    SVHN 
    &       & 10    &       &       & 26,032 \\
    Sun397 
    &       & 397   &       &       & 21,750 \\
    \midrule
    Patch Camelyon 
    & \multirow{4}[1]{*}{Specialized} & 2     & \multirow{4}[1]{*}{800/1,000} & \multirow{4}[1]{*}{200} & 32,768 \\
    EuroSAT 
    &       & 10    &       &       & 5,400 \\
    Resisc45 
    &       & 45    &       &       & 6,300 \\
    Retinopathy 
    &       & 5     &       &       & 42,670 \\
    \midrule
    Clevr/count 
    & \multirow{8}[1]{*}{Structured} & 8     & \multirow{8}[1]{*}{800/1,000} & \multirow{8}[1]{*}{200} & 15,000 \\
    Clevr/distance 
    &       & 6     &       &       & 15,000 \\
    DMLab 
    &       & 6     &       &       & 22,735 \\
    KITTI/distance 
    &       & 4     &       &       & 711 \\
    dSprites/location 
    &       & 16    &       &       & 73,728 \\
    dSprites/orientation 
    &       & 16    &       &       & 73,728 \\
    SmallNORB/azimuth 
    &       & 18    &       &       & 12,150 \\
    SmallNORB/elevation 
    &       & 9     &       &       & 12,150 \\
    \bottomrule
    \end{tabular}%

  \label{tab:dataset_vtab}
\end{table*}%

\section{Experimental Details on Large-Scale, Huge-Scale, and Hierarchical ViT Backbones}
\label{sec:LHSdetails}
In this section, we present the comprehensive results of our comparison among ViT-Large, ViT-Huge, and Swin-Base models, as discussed in Section \ref{sec:exp}. The detailed findings for each model are displayed in ~\cref{tab:vitL_vtab},~\ref{tab:vitH_vtab}, and \ref{tab:Swindetail}, respectively.

~\cref{tab:vitL_vtab} showcases the performance of the ViT-Large backbone across various tasks within the VTAB-1k benchmark. This table provides a thorough breakdown of the results, allowing for a detailed analysis of the model's strengths and weaknesses in different contexts.

Similarly, ~\cref{tab:vitH_vtab} presents the results for the ViT-Huge backbone. By comparing these results to those of the ViT-Large model, we can gain insights into the benefits and trade-offs associated with scaling up the model size.

Lastly, ~\cref{tab:Swindetail} outlines the performance of the Swin-Base model, which adopts a hierarchical architecture. This table enables us to assess the effectiveness of hierarchical designs in comparison to the standard ViT architectures.

By examining these detailed results, we can draw meaningful conclusions regarding the performance characteristics of large-scale, huge-scale, and hierarchical ViT backbones. These insights contribute to a deeper understanding of the capabilities and limitations of these models in various visual adaptation tasks.

\begin{table*}[!tbhp]
  \centering
  \caption{This table is extended from ~\cref{vitlhs_vtab} in Section~\ref{sec:exp} and describes the detailed experimental results of the performance comparison on VTAB-1k using ViT-Large pre-trained on ImageNet-21k as the backbone. Where Full fine-tuning and Linear probing are only used as controls and are not included in the bold comparison. The best results are shown in \textbf{bold}.}
  \resizebox{\linewidth}{!}{
    \begin{tabular}{c|ccccccc|c|cccc|c|cccccccc|c|cc}
    \bottomrule
    \multirow{2}[2]{*}{\diagbox{\textbf{Methods}}{\textbf{Datasets}}} & \multicolumn{8}{c|}{\textbf{Natural}}                         & \multicolumn{5}{c|}{\textbf{Specialized}} & \multicolumn{9}{c|}{\textbf{Structed}}                                &       &  \\
          & \rotatebox{90}{\textbf{CIFAR-100}} & \rotatebox{90}{\textbf{Caltech101}} & \rotatebox{90}{\textbf{DTD}} & \rotatebox{90}{\textbf{Flowers102}} & \rotatebox{90}{\textbf{Pets}} & \rotatebox{90}{\textbf{SVNH}} & \multicolumn{1}{c}{\rotatebox{90}{\textbf{Sun397}}} & \rotatebox{90}{\textbf{Mean}} & \rotatebox{90}{\textbf{Camelyon}} & \rotatebox{90}{\textbf{EuroSAT}} & \rotatebox{90}{\textbf{Resisc45}} & \multicolumn{1}{c}{\rotatebox{90}{\textbf{Retinopathy}}} & \rotatebox{90}{\textbf{Mean}} & \rotatebox{90}{\textbf{Clevr-Count}} & \rotatebox{90}{\textbf{Clevr-Dist}} & \rotatebox{90}{\textbf{DMLab}} & \rotatebox{90}{\textbf{KITTI-Dist}} & \rotatebox{90}{\textbf{dSpr-Loc}} & \rotatebox{90}{\textbf{dSpr-Ori}} & \rotatebox{90}{\textbf{sNORB-Azim}} & \multicolumn{1}{c}{\rotatebox{90}{\textbf{sNORB-Ele}}} & \rotatebox{90}{\textbf{Mean}} & \rotatebox{90}{\textbf{Mean Total}} & \rotatebox{90}{\textbf{Params.(M)}} \\
    \midrule
    Full fine-tuning & 68.6  & 84.3  & 58.6  & 96.3  & 86.5  & 87.5  & 41.4  & 74.7  & 82.6  & 95.9 & 82.4  & 74.2  & 83.8  & 55.4  & 55.0  & 42.2  & 74.2  & 56.8  & 43.0  & 28.5  & 29.7  & 48.1  & 65.4  & 303.4 \\
    Linear probing & 72.2  & 86.4  & 63.6  & 97.4  & 85.8  & 38.1  & 52.5  & 70.9  & 76.9  & 87.3  & 66.6  & 45.4  & 69.1  & 28.2  & 28.0  & 34.7  & 54.0  & 10.6  & 14.2  & 14.6  & 21.9  & 25.8  & 51.5  & 0.05 \\
			\midrule
			Adapter~\cite{houlsby2019parameter} & 75.3  & 84.2  & 54.5  & \textbf{97.4}  & \textbf{84.3}  & 31.3  &  \textbf{52.9}  &  68.6  &  75.8  &  85.1  &  63.4  &  69.5  &  73.5  &  35.4  &  \textbf{34.1}  &  30.8  &  47.1  &  30.4  &  23.4  &  10.8  &  19.8  &  29.0  &  52.9  & 2.38 \\
                Adapter+AOFT*  & \textbf{79.6}  & \textbf{89.6}  &  \textbf{63.0} & 84.3  & 73.7  & \textbf{72.2}  &  22.2  &  \textbf{70.7}  &  \textbf{77.1}  &  \textbf{86.2}  &  \textbf{71.2}  &  \textbf{73.6}  &  \textbf{77.0}  &  \textbf{68.2}  &  25.4  &  \textbf{39.3}  &  \textbf{66.8}  &  \textbf{61.3}  &  \textbf{42.4}  &  \textbf{29.9}  &  \textbf{21.8}  &  \textbf{44.4}  & \textbf{60.9}  & 0.10  \\
                \midrule
			LoRA~\cite{hu2021lora}  & 75.8  & 89.9  & 73.6  & 99.1  & 90.8  & \textbf{83.2}  &  57.5  &  81.4  &  86.0  &  95.0  &  83.4  &  \textbf{75.5}  &  85.0  &  78.1  &  60.5  &  46.7  &  \textbf{81.6}  &  76.7  &  51.3  &  28.0  &  35.4  &  57.3  &  72.0  & 0.74 \\
			LoRA+AOFT*   & \textbf{78.2}  & \textbf{95.0}  & \textbf{74.7}  & \textbf{99.5}  & \textbf{92.0}  & 82.4  &  \textbf{59.2}  &  \textbf{83.3}  &  \textbf{86.7}  &  \textbf{95.1}  &  \textbf{86.0}  &  75.2  &  \textbf{85.9}  &  \textbf{81.5}  &  \textbf{63.2}  &  \textbf{50.7}  &  81.0  &  \textbf{86.7}  &  \textbf{53.0}  &  \textbf{28.8}  &  \textbf{43.3}  &  \textbf{60.2}  &  \textbf{74.3}  & 0.15 \\
             
    \bottomrule
    \end{tabular}%
    }
  \label{tab:vitL_vtab}
\end{table*}

\begin{table*}[!tbhp]
  \centering
  \caption{This table is extended from ~\cref{vitlhs_vtab} in Section~\ref{sec:exp} and describes the detailed experimental results of the performance comparison on VTAB-1k using ViT-Huge pre-trained on ImageNet-21k as the backbone.  Where Full fine-tuning and Linear probing are only used as controls and are not included in the bold comparison. The best results are shown in \textbf{bold}.}
  \resizebox{\linewidth}{!}{
        \begin{tabular}{c|ccccccc|c|cccc|c|cccccccc|c|cc}
    \toprule
    \multirow{2}[2]{*}{\diagbox{\textbf{Methods}}{\textbf{Datasets}}} & \multicolumn{8}{c|}{\textbf{Natural}}    & \multicolumn{5}{c|}{\textbf{Specialized}} & \multicolumn{9}{c|}{\textbf{Structed}}                                &       &  \\
          & \rotatebox{90}{\textbf{CIFAR-100}} & \rotatebox{90}{\textbf{Caltech101}} & \rotatebox{90}{\textbf{DTD}} & \rotatebox{90}{\textbf{Flowers102}} & \rotatebox{90}{\textbf{Pets}} & \rotatebox{90}{\textbf{SVNH}} & \multicolumn{1}{c}{\rotatebox{90}{\textbf{Sun397}}} & \rotatebox{90}{\textbf{Mean}} & \rotatebox{90}{\textbf{Camelyon}} & \rotatebox{90}{\textbf{EuroSAT}} & \rotatebox{90}{\textbf{Resisc45}} & \multicolumn{1}{c}{\rotatebox{90}{\textbf{Retinopathy}}} & \rotatebox{90}{\textbf{Mean}} & \rotatebox{90}{\textbf{Clevr-Count}} & \rotatebox{90}{\textbf{Clevr-Dist}} & \rotatebox{90}{\textbf{DMLab}} & \rotatebox{90}{\textbf{KITTI-Dist}} & \rotatebox{90}{\textbf{dSpr-Loc}} & \rotatebox{90}{\textbf{dSpr-Ori}} & \rotatebox{90}{\textbf{sNORB-Azim}} & \multicolumn{1}{c}{\rotatebox{90}{\textbf{sNORB-Ele}}} & \rotatebox{90}{\textbf{Mean}} & \rotatebox{90}{\textbf{Mean Total}} & \rotatebox{90}{\textbf{Params.(M)}} \\
    \midrule
    Full fine-tuning & 58.7  & 86.5  & 55.0  & 96.5  & 79.7  & 87.5  & 32.5  & 70.9  & 83.1  & 95.5  & 81.9  & 73.8  & 83.6  & 47.6  & 53.9  & 37.8  & 69.9  & 53.8  & 48.6  & 30.2  & 25.8  & 46.0  & 63.1  & 630.90  \\
    Linear probing & 64.3  & 83.6  & 65.2  & 96.2  & 83.5  & 39.8  & 43.0  & 67.9  & 78.0  & 90.5  & 73.9  & 73.4  & 79.0  & 25.6  & 24.5  & 34.8  & 59.0  & 9.5   & 15.6  & 17.4  & 22.8  & 26.1  & 52.7  & 0.06  \\
    \midrule
			Adapter~\cite{houlsby2019parameter} & \textbf{69.4}  & 84.4  & 62.7  & 97.2  & 84.2  & 33.6  & 45.3  & 68.1  & 77.3  & 86.6  & 70.8  & 71.1  & 76.4  & 28.6  & 27.5  & 29.2  & 55.2  & 10.0  & 15.2  & 11.9  & 18.6  & 24.5  & 51.5  & 5.78  \\
                Adapter+AOFT*  & 67.9  & \textbf{91.3}  &  \textbf{69.6} & \textbf{98.6}  & \textbf{88.1}  & \textbf{79.4}  &  \textbf{49.0}  & \textbf{77.7}   &  \textbf{80.0}  &  \textbf{95.3}  &  \textbf{78.3}  &  \textbf{73.6}  & \textbf{81.8}   &  \textbf{31.5}  & \textbf{31.7}   &  \textbf{39.0}  &  \textbf{71.6}  &  \textbf{40.2}  &  \textbf{22.4}  &  \textbf{23.8}  &  \textbf{36.9}  & \textbf{37.1}   & \textbf{61.5}  & 0.17  \\
        \midrule
			LoRA~\cite{hu2021lora} & 63.0  & 89.4  & 68.1  & 98.0  & 87.0  & \textbf{85.2}  & 48.7  & 77.1  & 82.2  & 94.3  & \textbf{83.1}  & 74.2 & 83.5  & 68.6  & \textbf{65.0}  & 44.8  & 76.4 & 70.8  & \textbf{48.8}  & 30.4  & 38.3 & 55.4 & 69.3  & 1.21  \\
			LoRA+AOFT*   & \textbf{68.9}  & \textbf{93.0}  & \textbf{69.9}  & \textbf{98.7}  & \textbf{89.1}  & 80.9  &  \textbf{51.5}  &  \textbf{78.8}  &  \textbf{84.2}  &  \textbf{94.5}  &  82.1  &  \textbf{74.6}  &  \textbf{83.8}  &  \textbf{74.1}  &  63.5  &  \textbf{46.5}  &  \textbf{79.3}  &  \textbf{79.9}  &  48.7  &  \textbf{31.5}  &  \textbf{43.2}  &  \textbf{58.3}  &  \textbf{71.3}  & 0.20 \\
            
			\bottomrule
    \end{tabular}%
    }

  \label{tab:vitH_vtab}
\end{table*}

\begin{table*}[!tbhp]
  \centering
    \caption{This table is extended from ~\cref{tab:swinb_vtab} in Section~\ref{sec:exp} and describes the detailed experimental results of the performance comparison on VTAB-1k using Swin-Base pre-trained on ImageNet-21k as the backbone. The best results are shown in \textbf{bold}.}
  \resizebox{\linewidth}{!}{
    \begin{tabular}{c|ccccccc|c|cccc|c|cccccccc|c|cc}
    \toprule
    \multirow{2}[2]{*}{\diagbox{\textbf{Methods}}{\textbf{Datasets}}} & \multicolumn{8}{c|}{\textbf{Natural}}                         & \multicolumn{5}{c|}{\textbf{Specialized}} & \multicolumn{9}{c|}{\textbf{Structed}}                                &       &  \\
          & \rotatebox{90}{\textbf{CIFAR-100}} & \rotatebox{90}{\textbf{Caltech101}} & \rotatebox{90}{\textbf{DTD}} & \rotatebox{90}{\textbf{Flowers102}} & \rotatebox{90}{\textbf{Pets}} & \rotatebox{90}{\textbf{SVNH}} & \multicolumn{1}{c}{\rotatebox{90}{\textbf{Sun397}}} & \rotatebox{90}{\textbf{Mean}} & \rotatebox{90}{\textbf{Camelyon}} & \rotatebox{90}{\textbf{EuroSAT}} & \rotatebox{90}{\textbf{Resisc45}} & \multicolumn{1}{c}{\rotatebox{90}{\textbf{Retinopathy}}} & \rotatebox{90}{\textbf{Mean}} & \rotatebox{90}{\textbf{Clevr-Count}} & \rotatebox{90}{\textbf{Clevr-Dist}} & \rotatebox{90}{\textbf{DMLab}} & \rotatebox{90}{\textbf{KITTI-Dist}} & \rotatebox{90}{\textbf{dSpr-Loc}} & \rotatebox{90}{\textbf{dSpr-Ori}} & \rotatebox{90}{\textbf{sNORB-Azim}} & \multicolumn{1}{c}{\rotatebox{90}{\textbf{sNORB-Ele}}} & \rotatebox{90}{\textbf{Mean}} & \rotatebox{90}{\textbf{Mean Total}} & \rotatebox{90}{\textbf{Params.(M)}} \\
    \midrule
    Full fine-tuning & 72.2  & 88.0  & 71.4  & 98.3  & 89.5  & 89.4  & 45.1  & 79.1  & 86.6  & \textbf{96.9} & \textbf{87.7} & 73.6  & 86.2  & 75.7 & \textbf{59.8} & 54.6 & 78.6  & 79.4  & 53.6 & \textbf{34.6} & \textbf{40.9} & 59.7 & 72.4  & 86.9 \\
    Linear probing & 61.4  & 90.2  & 74.8  & 95.5  & 90.2  & 46.9  & \textbf{55.8} & 73.5  & 81.5  & 90.1  & 82.1  & 69.4  & 80.8  & 39.1  & 35.9  & 40.1  & 65.0  & 20.3  & 26.0  & 14.3  & 27.6  & 33.5  & 58.2  & 0.05 \\
    \midrule
    MLP-4~\cite{jia2022visual} & 54.9  & 87.4  & 71.4  & 99.5  & 89.1  & 39.7  & 52.5  & 70.6  & 80.5  & 90.9  & 76.8  & 74.4  & 80.7  & 60.9  & 38.8  & 40.2  & 66.5  & 9.4   & 21.1  & 14.5  & 28.8  & 31.2  & 57.7  & 4.04 \\
    Partial~\cite{jia2022visual} & 60.3  & 88.9  & 72.6  & 98.7  & 89.3  & 50.5  & 51.5  & 73.1  & 82.8  & 91.7  & 80.1  & 72.3  & 81.7  & 34.3  & 35.5  & 43.2  & 77.1  & 15.8  & 26.2  & 19.1  & 28.4  & 35.0  & 58.9  & 12.65 \\
    Bias~\cite{zaken2022bitfit}  & 73.1  & 86.8  & 65.7  & 97.7  & 87.5  & 56.4  & 52.3  & 74.2  & 80.4  & 91.6  & 76.1  & 72.5  & 80.1  & 47.3  & 48.5  & 34.7  & 66.3  & 57.6  & 36.2  & 17.2  & 31.6  & 42.4  & 62.1  & 0.25 \\
    VPT-Shallow~\cite{jia2022visual} & 78.0 & 91.3 & 77.2 & 99.4 & 90.4  & 68.4  & 54.3  & 79.9  & 80.1  & 93.9  & 83.0  & 72.7  & 82.5  & 40.8  & 43.9  & 34.1  & 63.2  & 28.4  & 44.5  & 21.5  & 26.3  & 37.8  & 62.9  & 0.05 \\
    VPT-Deep~\cite{jia2022visual} & \textbf{79.6} & 90.8 & \textbf{78.0} & 99.5 & 91.4 & 46.5  & 51.7  & 76.8  & 84.9  & 96.2 & 85.0  & 72.0  & 84.5  & 67.6  & 59.4 & 50.1  & 74.1  & 74.4  & 50.6  & 25.7  & 25.7  & 53.4  & 67.7  & 0.22 \\
    ARC~\cite{dong2023efficient}   & 62.5  & 90.0  & 71.9  & 99.2  & 87.8  & 90.7 & 51.1  & 79.0 & \textbf{89.1} & 95.8  & 84.5  & 77.0 & 86.6 & 75.4 & 57.4  & 53.4  & 83.1 & \textbf{91.7} & \textbf{55.2} & 31.6 & 31.8  & 59.9 & 72.6 & 0.27 \\
    
    RLRR~\cite{dong2024low}  & 66.1  & 90.6  & 75.5  & 99.3  & 92.1 & \textbf{90.9} & 54.7 & 81.3 & 87.1 & 95.9  & 87.1 & 76.5 & 86.7 & 66.0  & 57.8  & \textbf{55.3} & 84.1 & 91.1 & \textbf{55.2} & 28.6  & 34.0 & 59.0  & 73.0 & 0.41 \\
    \midrule
    LoRA+AOFT*   & 71.8  & \textbf{92.3}  & 77.1  & \textbf{99.5}  & \textbf{92.6}  & 86.4  &  \textbf{55.8}  & \textbf{82.8}   &  86.9  &  96.4  &  87.3  &  \textbf{77.6}  &  \textbf{87.1}  &  \textbf{84.5}  &  59.3  &  53.6  &  \textbf{84.7}  &  86.8  &  52.3  &  28.1  &  35.5  & \textbf{60.6}   &  \textbf{73.3} & 0.14 \\
    \bottomrule
    \end{tabular}%
    }

  \label{tab:Swindetail}
\end{table*}

\section{Orthogonality of Operators}
\label{sec:operator_Q}
In this work, a method is applied to generate multiple orthogonal basis from a vector to form a column of orthogonal matrices. The matrix $\mathbf{Q}$ is derived from the matrix:

\begin{equation}
  \begin{bmatrix}
      \!\cos\varphi \!&\!\! -x_1\sin\varphi \!&\!\! \cdots \!&\!\! -x_i\sin\varphi \!\\
      \!x_1\sin\varphi \!&\!\! 1\!+\!x_1^2(\cos\varphi\!-\!1) \!&\!\! \cdots \!&\!\! x_ix_1(\cos\varphi\!-\!1) \!\\
      \!x_2\sin\varphi \!&\!\! x_1x_2(\cos\varphi\!-\!1) \!&\!\! \cdots \!&\!\! x_ix_2(\cos\varphi\!-\!1) \!\\
      \!\vdots \!&\!\! \vdots \!&\!\! \vdots \!&\!\! \vdots \!\\
      \!x_i\sin\varphi \!&\!\! x_1x_i(\cos\varphi\!-\!1) \!&\!\! \cdots \!&\!\!\! 1\!+\!x_i^2(\cos\varphi\!-\!1) \!\\
      \!\vdots \!&\!\! \vdots \!\!&\! \vdots \!&\! \vdots \!\\
      \!x_N\sin\varphi \!&\!\! x_1x_N(\cos\varphi\!-\!1) \!&\!\! \cdots \!&\!\!\! x_ix_N(\cos\varphi\!-\!1)\!
  \end{bmatrix}\!,
  \label{eq:cos matricex}
\end{equation}

\noindent
let $\cos\varphi = q_0$ and $\sin\varphi=(\sum_{i=1}^N|q_i|^2)^{1/2}$, where $\sum_{i=1}^N|q_i|^2=1$, that means $x_i=\frac{q_i}{\sin\varphi}(i=1,2,...,N)$, $x_ix_j(\cos\varphi-1)$ can be calculated by this way as follows:
\begin{equation}
    \begin{split} 
    x_jx_i(\cos\varphi-1) & =\frac{q_j}{\sin\varphi}\frac{q_i}{\sin\varphi}(\cos\varphi-1) \\
    & =\frac{q_jq_i}{\sin^2\varphi}(\cos\varphi-1) \\
    & =-\frac{q_jq_i}{1+\cos\varphi}=-\frac{q_jq_i}{1+q_0}.
    \end{split}
    \label{eq:simplify}
\end{equation}
Based on \cref{eq:cos matricex} and \cref{eq:simplify}, we obtain the simplified result as \cref{Eq:AOFT}.

\section{Detailed Configuration}
\cref{tab:optimizer_details} summarizes the detailed configurations we used for experiments. As mentioned in Section~\ref{sec:exp}, we utilize grid search to select hyper-parameters such as learning rate, weight decay, batch size, and dropout rate, using the validation set of each task. 

\begin{table*}[t]
  \centering
  \caption{The implementation details of configurations such as optimizer and hyper-parameters. We select the best hyper-parameters for each download task via using grid search.}
  \label{tab:optimizer_details}%
    \begin{tabular}{c|c}
    \toprule
    Optimizer & AdamW \\
    \midrule
    Learning Rate & \{0.2, 0.1, 0.05, 0.01, 0.005, 0.001, 0.0001\} \\
    Weight Decay & \{0.05, 0.01, 0.005, 0.001, 0\} \\
    Dropout Rate & \{0, 0.1, 0.3, 0.5, 0.7\} \\
    Batch Size & \{256, 128, 32\} \\
    Learning Rate Schedule & Cosine Decay \\
    Training Epochs  & 100 \\
    Warmup Epochs & 10 \\
    \bottomrule
    \end{tabular}%
  
\end{table*}

\begin{table*}[!tbhp]
	\centering
        \caption{Based on the ViT-B backbone, a comparison of the results is presented for applying AOFT to both the FFN and MHA layers, as well as with or without incorporating eigenvalues, on the VTAB dataset.}
	\resizebox{\linewidth}{!}{
		\begin{tabular}{c|ccccccc|c|cccc|c|cccccccc|c|cc}
			\toprule
			\multirow{2}[2]{*}{\diagbox{\textbf{Methods}}{\textbf{\rotatebox{0}{Datasets}}}} & \multicolumn{8}{c|}{\textbf{Natural}}                         & \multicolumn{5}{c|}{\textbf{Specialized}} & \multicolumn{9}{c|}{\textbf{Structed}}                                &       &  \\
			& \rotatebox{90}{\textbf{CIFAR-100}} & \rotatebox{90}{\textbf{Caltech101}} & \rotatebox{90}{\textbf{DTD}} & \rotatebox{90}{\textbf{Flowers102}} & \rotatebox{90}{\textbf{Pets}} & \rotatebox{90}{\textbf{SVNH}} & \multicolumn{1}{c}{\rotatebox{90}{\textbf{Sun397}}} & \rotatebox{90}{\textbf{Mean}} & \rotatebox{90}{\textbf{Camelyon}} & \rotatebox{90}{\textbf{EuroSAT}} & \rotatebox{90}{\textbf{Resisc45}} & \multicolumn{1}{c}{\rotatebox{90}{\textbf{Retinopathy}}} & \rotatebox{90}{\textbf{Mean}} & \rotatebox{90}{\textbf{Clevr-Count}} & \rotatebox{90}{\textbf{Clevr-Dist}} & \rotatebox{90}{\textbf{DMLab}} & \rotatebox{90}{\textbf{KITTI-Dist}} & \rotatebox{90}{\textbf{dSpr-Loc}} & \rotatebox{90}{\textbf{dSpr-Ori}} & \rotatebox{90}{\textbf{sNORB-Azim}} & \multicolumn{1}{c}{\rotatebox{90}{\textbf{sNORB-Ele}}} & \rotatebox{90}{\textbf{Mean}} & \rotatebox{90}{\textbf{Mean Total}} & \rotatebox{90}{\textbf{Params.(M)}} \\
            \midrule
            LoRA$(\mathbf{W}_q,\mathbf{W}_v)$ & 67.1 & 91.4 & 69.4 & 98.9  & 90.4 & 85.3 & 54.0 & 79.5 & 84.9 & 95.3 & 84.4 & 73.6 & 84.6 & 82.9 & 69.2 & 49.8 & 78.5 & 75.7 & 47.1 & 31.0 & 44.0 & 59.8 & 72.3 & 0.29 \\
            LoRA+AOFT$(\mathbf{W}_q,\mathbf{W}_v)$ & 74.0 & 91.0 & 72.7 & 99.3 & 89.3 & 80.6 & 56.8 & 80.52 & 84.9 & 94.6 & 82.7 & 75.6 & 84.4 & 71.4 & 57.5 & 42.7 & 82.0 & 83.4 & 53.9 & 22.6 & 44.5 & 57.3 & 71.5 & 0.08 \\
            LoRA+AOFT$(\mathbf{W}_q,\mathbf{W}_v,\mathbf{W}_\mathrm{FC1},\mathbf{W}_\mathrm{FC2})$ & 75.0 & 93.2 & 70.2 & 99.2 & 91.1 & 84.0 & 57.4 & 81.7 & 85.1 & 95.6 & 84.7 & 75.6 & 85.3 & 79.8 & 60.5 & 49.1 & 82.0 & 79.64 & 54.92 & 32.7 & 45.8 &  60.6 &  73.4 & 0.16  \\
            \midrule
            Adapter$(\mathbf{W}_\mathrm{FFN})$ & 69.2  & 90.1  & 68.0  & 98.8  & 89.9  & 82.8  &  54.3  & 79.0   &  84.0  &  94.9  &  81.9  &  75.5  &  84.1  &  80.9 &  65.3  &  48.6  &  78.3  &  74.8  &  48.5  &  29.9  &  41.6  & 58.5   &  71.4  &  0.16 \\
            Adapter+AOFT$(\mathbf{W}_\mathrm{FFN})$ & 74.1 & 93.9 & 72.6 & 99.4 & 91.0 & 82.9 & 57.6 & 79.0 & 85.8 & 95.1 & 83.4 & 76.3 & 84.1 & 77.7 & 61.3 & 49.1 & 80.0 & 80.8 & 53.8 & 30.4 & 42.8 & 58.5 & 71.4 & 0.16 \\
            Adapter+AOFT$(\mathbf{W}_\mathrm{FFN},\mathbf{W}_\mathrm{MHA})$ & 67.6 & 93.9 & 72.1 & 99.3 & 91.6 & 86.4 & 53.8 & 80.7 &  86.8 & 95.5 & 86.0	 & 76.5	  & 86.2 &  79.0 & 62.6 & 51.5 & 82.7 & 87.0 & 54.8 & 27.7 & 42.5 & 60.9 & 73.5 & 0.08 \\
		\bottomrule
		\end{tabular}
	}
	\label{Tab:vitb_vtab_adapter_ablation}
\end{table*}

\section{Experimental details on ablation study}
In addition to incorporating our method into the MHA and FFN layers, we also added AOFT solely to the MHA layer. In order to better compare the effects of AOFT in the adapter method across different layers, we conducted the following experiments, \cref{Tab:vitb_vtab_adapter_ablation} shows the experimental results.

\clearpage
\end{appendix}

\end{document}